\documentclass[10pt,twocolumn,letterpaper]{article}

\usepackage{iccv}              

\usepackage{tcolorbox}
\usepackage{float}
\usepackage{colortbl}

\newcommand{\ours}{Hybrid-depth}

\definecolor{linecolor1}{RGB}{246, 248, 239}
\definecolor{linecolor2}{RGB}{230, 234, 217}
\definecolor{linecolor3}{RGB}{211, 222, 190}

\newcounter{question}
\newcommand{\question}{%
    \stepcounter{question}%
    \noindent\textbf{Q\thequestion:~\ignorespaces}%
}

\definecolor{iccvblue}{rgb}{0.21,0.49,0.74}
\usepackage[pagebackref,breaklinks,colorlinks,allcolors=iccvblue]{hyperref}

\definecolor{barrier}{RGB}{112,128,144}

\definecolor{runpei-orange}{HTML}{F35F27}

\def\paperID{1976} 
\def\confName{ICCV}
\def\confYear{2025}

\title{Hybrid-grained Feature Aggregation with Coarse-to-fine Language Guidance for Self-supervised Monocular Depth Estimation}

\vspace{-15mm}

\author{
  Wenyao Zhang$^{123}$\thanks{Equal contribution. $^\ddagger$Corresponding author.} \qquad Hongsi Liu$^{234*}$ \qquad Bohan Li$^{123*}$ \qquad Jiawei He$^{5}$ \qquad Zekun Qi$^{6}$ \\
  Yunnan Wang$^{123}$ \quad Shengyang Zhao$^{23}$\qquad Xinqiang Yu$^{5}$ \qquad Wenjun Zeng$^{23}$ \qquad Xin Jin$^{23\ddagger}$ \vspace{5pt} \\
  $^1$MoE Key Lab of Artificial Intelligence, AI Institute, Shanghai Jiao Tong University \\
  $^2$Ningbo Institute of Digital Twin, Eastern Institute of Technology, Ningbo, China \\
  $^3$Ningbo Key Laboratory of Spatial Intelligence and Digital Derivative, Ningbo, China \\
  $^4$University of Science and Technology of China \quad \quad $^5$CASIA \quad \quad $^6$Tsinghua University
}

\begin{document}
\maketitle


\begin{abstract}
Current self-supervised monocular depth estimation (MDE) approaches encounter performance limitations due to insufficient semantic-spatial knowledge extraction. To address this challenge, we propose Hybrid-depth, a novel framework that systematically integrates foundation models (\eg, CLIP and DINO) to extract visual priors and acquire sufficient contextual information for MDE. Our approach introduces a coarse-to-fine progressive learning framework: 
1) Firstly, we aggregate multi-grained features from CLIP (global semantics) and DINO (local spatial details) under contrastive language guidance. A proxy task comparing close-distant image patches is designed to enforce depth-aware feature alignment using text prompts;
2) Next, building on the coarse features, we integrate camera pose information and pixel-wise language alignment to refine depth predictions. This module seamlessly integrates with existing self-supervised MDE pipelines (\eg, Monodepth2, ManyDepth) as a plug-and-play depth encoder, enhancing continuous depth estimation.
By aggregating CLIP’s semantic context and DINO’s spatial details through language guidance, our method effectively addresses feature granularity mismatches.
Extensive experiments on the KITTI benchmark demonstrate that our method significantly outperforms SOTA methods across all metrics, which also indeed benefits downstream tasks like BEV perception. Code is available at {\small\url{https://github.com/Zhangwenyao1/Hybrid-depth}}.
\end{abstract}
\vspace{-3mm}

\section{Introduction}
\label{sec:intro}
\begin{figure}[htbp]
    \centering
    \includegraphics[width=1\linewidth]{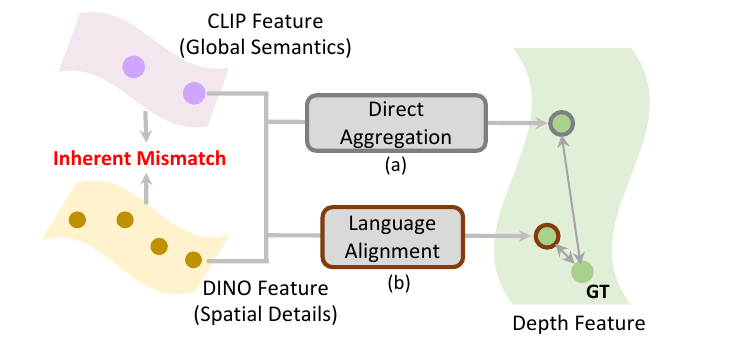}
    \vspace{-8mm}
    \caption{CLIP and DINO exhibit complementary strengths: CLIP excels in capturing global semantic context, while DINO specializes in local spatial detail extraction. However, their fusion is hindered by inherent feature-level mismatches. Direct aggregation strategies like channel concatenation (Fig. 1(a)) result in suboptimal depth representations due to misaligned semantic and spatial features. In contrast, our approach (Fig. 1(b)) employs depth-aware language prompts as a granularity calibrator to align cross-level features into a unified depth hierarchy, ensuring semantic coherence and spatial precision.\looseness=-1
    }
    \vspace{-6mm}
    \label{fig:enter-label}
\end{figure}

\begin{figure*}[tb]
\vspace{-2mm}
  \centering
  \includegraphics[width=0.95\linewidth]{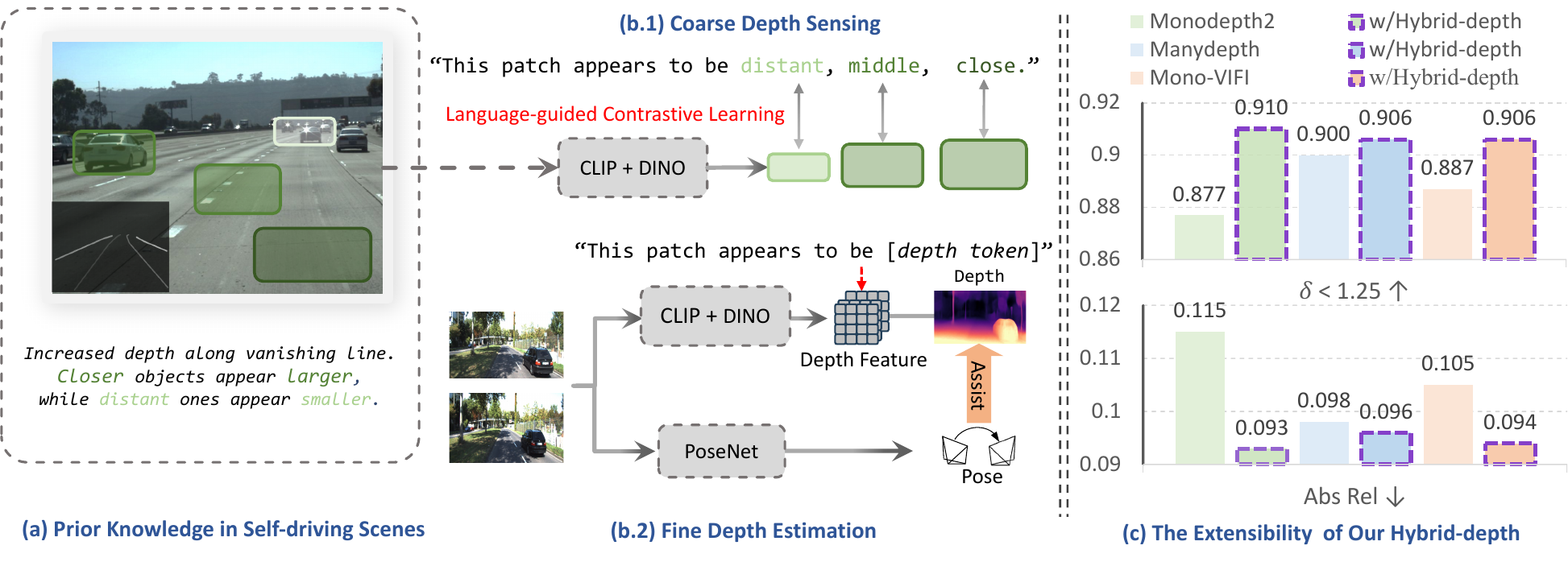}
  \vspace{-5mm}
  \caption{
    The detailed pipeline of our proposed method. We first aggregate different-grained features from CLIP and DINO for coarse depth sensing under 
    contrastive language guidance (Fig.~2(b.1)), incorporating prior geometric knowledge (Fig.~2(a)).
    These models are then optimized with the help of the auxiliary camera pose from PoseNet like existing methods to learn a fine depth estimation (Fig.~2(b.2)). 
    It also can be extended to improve the capture of continuous depth variations.
    By equipping our method, existing self-supervised MDE methods (\eg, Monodepth2~\cite{monodepthv2}, ManyDepth~\cite{manydepth21}, and Mono-VIFI~\cite{liu2024mono}) achieve significant performance improvements (Fig.~2(c)). \looseness=-1
    }
    \vspace{-5mm}
  \label{fig: teaser}
\end{figure*}



Monocular depth estimation (MDE) is a challenging task involving the accurate prediction of per-pixel depth values within a scene from a single image. It plays a critical role  in advancing 3D vision, especially in applications such as autonomous driving \cite{wang2019pseudo, li2024uniscene,you2020pseudo,li2024one,okae2021robust}, robotics~\cite{wofk2019fastdepth, zhuhu, sofar25,zhang2025dreamvla,zhang2022predict,zhang2022analysis}, and 3D reconstruction~\cite{du2020depthlab}.
Recent advancements in deep learning methodologies have enabled supervised MDE algorithms~\cite{adabins,depthanything,bhat2023zoedepth,sharpdepth25} to achieve notable improvements in depth estimation accuracy. Nonetheless, the reliance on large-scale datasets with per-pixel depth annotations for supervised training remains a significant bottleneck, limiting their practical applicability in realistic scenarios.

To address these limitations, numerous studies~\cite{zhou2017, monodepthv2, struct2depth, bian2021unsupervised} have explored self-supervised MDE by exploiting 3D geometric consistency in monocular video sequences. For instance, existing methods typically employ separate network branches to predict depth and camera pose~\cite{monodepthv2, zhou2017}, optimizing the model via image reconstruction and smoothness losses. However, this paradigm suffers from constrained data diversity and limited model capacity, resulting in performance gaps compared to supervised approaches.

The remarkable success of foundation models such as CLIP~\cite{clip} and DINO~\cite{dinov121,dinov223}, pre-trained on massive datasets and subsequently fine-tuned for downstream tasks, has revolutionized a range of vision applications including zero-/few-shot classification~\cite{COOP,coco,maple}, detection~\cite{zhong2022regionclip,gu2021open}, and segmentation~\cite{denseclip,dong2023maskclip,lueddecke22cvpr}.
This raises the question: ``Can large-scale pre-training on coarse data, followed by task-specific fine-tuning, advance self-supervised MDE ?''
Recent works~\cite{tong2024eyes,fazli2024clip,openvla} highlight the complementary strengths of CLIP and DINO: CLIP captures global semantic context (\eg, object categories, scene layouts) through text-image alignment, while DINO excels at local spatial detail extraction (\eg, edges, local geometry) via self-supervised patch-level learning. 
This synergy suggests a promising pathway for improving self-supervised MDE.
However, as shown in Fig.~\ref{fig:enter-label}, naive feature fusion (\eg, direct concatenation) often produces suboptimal results due to inherent feature-level mismatches. 
Specifically, CLIP’s global semantic representations lack spatial precision for depth estimation, whereas DINO’s local features fail to encode contextual depth hierarchies~\cite{10891864, sparse25, cliptodino23}. Consequently, harmonizing these complementary capabilities remains a pivotal challenge for advancing self-supervised MDE. \looseness=-1

\begin{figure*}[htbp]
  \centering
  \includegraphics[width=1.0\linewidth]{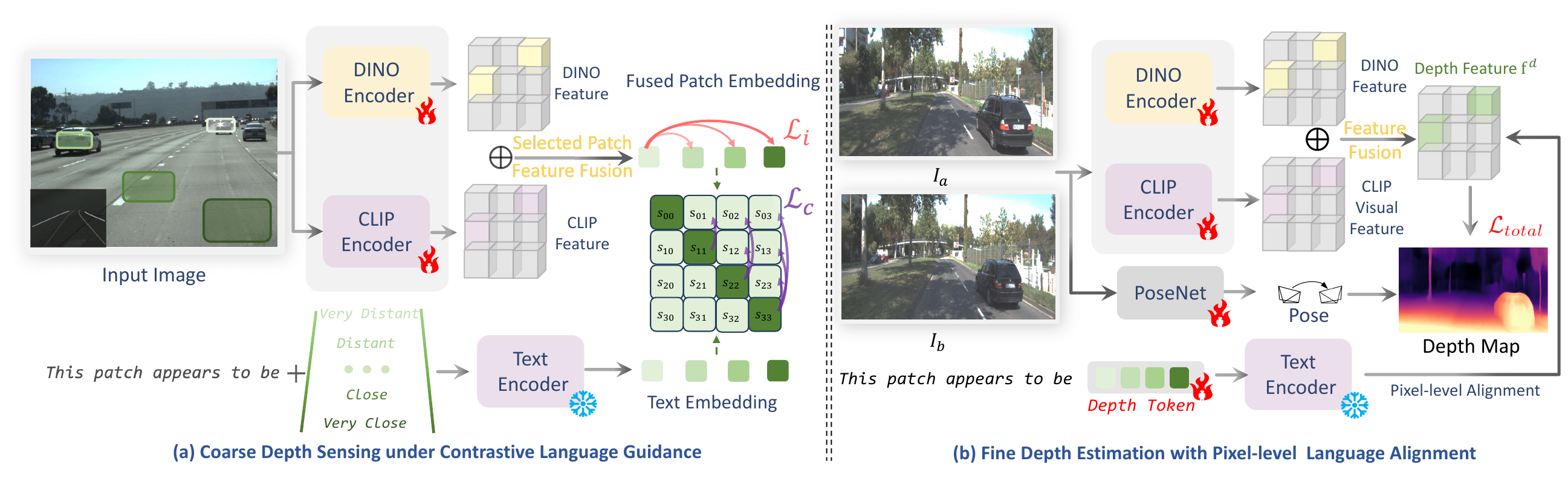}
  \vspace{-7mm}
  \caption{Left: For the coarse depth sensing stage, we first aggregate the CLIP and DINO features and then design two contrastive learning strategies to endow them with coarse depth sensing capabilities by leveraging geometric priors from self-driving scenes.
  Right: During the fine depth estimation phrase, different from previous methods, \ours~ not only combines camera pose information from PoseNet but also conducts pixel-wise alignment with learnable depth prompts for hybrid-grained features to learn a fine depth estimation ability. \looseness=-1
  }
  \label{fig: method}
  \vspace{-5mm}
\end{figure*}

To address these issues, we propose Hybrid-depth, a novel framework that systematically integrates foundation models (\eg, CLIP and DINO) to extract visual priors and contextual information for MDE.
Our approach introduces a coarse-to-fine learning scheme that progressively endows CLIP and DINO with depth estimation capabilities while fully exploiting their hybrid-grained features through language alignment.
As illustrated in Fig.~\ref{fig: teaser}, 
\ours~consists of two phases: 
1). \textbf{Coarse depth sensing}. This stage aims to harness geometric priors from autonomous driving scenes (\eg, depth gradients along vanishing lines) to design a proxy task with close-distant patch comparison.
Specifically, we extract multi-grained features from CLIP and DINO along lane markings in images and align these feature patches with depth-related textual prompts  (\eg, “This patch appears to be [\textit{close/far/very far}]”) to formulate two novel contrastive learning losses.
Unlike previous 2D CLIP-based methods \cite{auty2023learning,depthclip,hu2024learning} that employ foundation models solely as depth encoders, this stage equips CLIP and DINO with coarse depth reasoning ability through structured semantic guidance.
2). \textbf{Fine depth estimation}. This stage refines Hybrid-depth to achieve precise depth estimation.
Different from previous methods, \ours~not only combines camera pose information but also conducts pixel-wise alignment with learnable depth prompts for hybrid-grained features.
Due to the modular independence from typical self-supervised MDE pipelines, Hybrid-depth can be seamlessly integrated as an enhanced depth encoder to boost performance in existing frameworks.
Additionally, the coarse-to-fine language guidance harmonizes multi-grained features by acting as a granularity calibrator (Fig.~\ref{fig:enter-label}), balancing the model’s focus between high-level semantic richness (from CLIP) and low-level spatial precision (from DINO). 
This structured multi-scale feature handling maximizes the complementary strengths of both models.
The key contributions of our work are summarized as follows: \looseness=-1

\begin{itemize}
     \item To the best of our knowledge, our framework is the first work that leverages foundation models (CLIP and DINO) for self-supervised MDE. Our approach introduces a coarse-to-fine learning framework that progressively transfers 2D priors from foundation models to enhance 3D geometric perception.\looseness=-1
    \item By synergistically aggregating CLIP’s semantic context and DINO’s spatial details through language guidance, we defectively resolve feature granularity mismatches.
    \item Our method functions as an efficient depth encoder compatible with existing self-supervised MDE pipelines, achieving significant performance improvements.
\end{itemize}


\section{Related Works}

\subsection{Monocular Depth Estimation}

\subsubsection{Supervised Monocular Depth Estimation}
Early researchers \cite{hoiem2007recovering,liu2008sift,make3d} conduct monocular depth estimation based on traditional computer vision techniques and hand-crafted features.
However, these methods encounter limitations due to their reliance on explicit depth cues.
With the development of deep learning, certain approaches can efficiently learn regressing depth from numerous annotations \cite{KITTI}.
In detail, Eigen \textit{et al.} \cite{eigen2014depth} first proposes a multi-scale fusion network to regress the depth value from the RGB images.
Based on this, many researches concentrate on improving the accuracy \cite{d3depth,Wang2022CVPR, kim2024clip,ecodepth24,surprising24, monodiffusion24, Park24, Ke24, soccernet24, depthpro24,rsa24, Piccinelli2024CVPR, khan2021differentiable,li2018megadepth} and generalization \cite{Nguyen2024CVPR,wei2024d,MAL24,Quat24,zeng2024priordiffusion}.
Furthermore, some works \cite{adabins,binsformer,fu2018deep} propose conducting depth regression as a classification task, 
and other methods choose to introduce more priors like segmentation \cite{shao2023nddepth}, geometry\cite{vnl}, and better objective functions \cite{redweb}. 
Recently, many studies \cite{midas22, bhat2023zoedepth, depthanything,depthanythingv2, metric3d23, metric3d24,wangdepth24, li2023bridging,okae2021robust,ElBanani2024CVPR,zou2025,guizilini2024grin} propose pre-trained models on numerous depth labels that would work well in generalizable metrics.
Nevertheless, collecting vast amounts of manually labeled data is resource-consuming.

\subsubsection{Self-supervised Monocular Depth Estimation}

To address the above limitations inherent in the supervised MDE, Zhou \textit{et al.}~\cite{zhou2017} introduces incorporating an additional PoseNet to learn the 6-DoF camera pose from monocular videos, enabling the prediction of depths.
Building on this foundation, many advanced techniques~\cite{monodepthv2, packnet-semguided, yin2018geonet,lyu2020hr,Walia2022CVPR,bian2021unsupervised,dualpathdepth25, dong2024ppea,cheng2024self,Chawla2024WACV, li2024hierarchical,10610318,wang2023planedepth,poggi2020uncertainty,Depthformer22,Si2023CVPR} are developed to boost the performance.
Other studies leverage supplementary supervisory information, such as traditional stereo matching ~\cite{watson2019self} and semantic cues~\cite{jiao2018look, klingner2020selfsupervised}.
Manydepth~\cite{manydepth21} proposes employing multiple frames available at test time and leveraging geometric constraints by constructing a cost volume, resulting in superior performance.
To tackle the well-known edge-fattening issue, Bello. \textit{et al.} \cite{gonzalezbello2020forget} and Zhu \textit{et al.} \cite{zhu2020edge} utilize an occlusion mask to remove the incorrect supervision under photometric loss.
Furthermore, some works investigate to estimate depths in more challenging environments like bad weather \cite{saunders2023self}, night time \cite{vankadari2020unsupervised} and indoor environments \cite{ji2021monoindoor, zhao2023gasmono}.
However, these methods typically encounter performance bottlenecks and overfitting dilemmas due to limited data.

\subsection{Foundation Models and Prompt Learning}
Recently, CLIP \cite{clip} and ALIGN \cite{align} leverage large-scale image-text datasets containing numerous samples to learn a joint representation by contrastive learning.
By mapping image and text information into a common space to compute the similarity of the text and image, the models can gain a more comprehensive understanding of visual and textual inputs.
Additionally, single-modal methods focus on learning from visual data alone: MAE~\cite{mae} reconstructs masked image patches to model spatial context, and MoCo~\cite{moco} builds dynamic dictionaries via momentum contrast to capture discriminative features. 
DINO ~\cite{dinov121,dinov223} uses self-distillation with multi-crop augmentation to learn hierarchical visual patterns.
Furthermore, these pre-trained foundation models also show remarkable transferability for novel datasets \cite{clip, align} and other downstream tasks~\cite{GLIP, zhu2023pointclip, zhang2022pointclip, denseclip, li2023clip, Luo2023SegCLIP, liang2023open, liang2023crowdclip, act23, YANG2024106410, shapellm24, recon23}.

Prompt learning is tailored to optimize text prompts to facilitate better downstream performance while maintaining the generalization of models.
Inspired by the recent advances \cite{petroni2019language, petroni2019language, prompttuning} in NLP, prompt learning has become prevalent in computer vision and multimodal models.
Zhou \textit{et al.}\cite{COOP, COCOOP} introduce prompt learning to CLIP, which changes the input text from the handcrafted template to learnable context vectors and gains significant improvements.
Following this, many advanced techniques \cite{maple, proda, maple, chen2022prompt, wenyao24,vpt} are proposed to boost performance.
Additionally, researchers investigate employing prompt learning to benefit different downstream tasks \cite{yao2021cpt,zhu2023segprompt,li2023clip}. 
However, there are few studies on self-supervised MDE using foundation models.

DepthCLIP~\cite{depthclip} is a pioneering attempt to conduct zero-shot depth estimation using CLIP, but the predicted depth is ambiguous.
Furthermore, Dylan \textit{et al.}~\cite{auty2023learning}, Hu \textit{et.al.}~\cite{hu2024learning} and Eunjin~\textit{et.al.} \cite{clipcabins24} reveal that learnable prompts outperform the human handcraft template, but they still underperform previous MDE methods.
It implies existing self-supervised MDE methods with CLIP do not fully unleash the power of the vision-language model and knowledge \cite{zeng2024wordepth}.
Therefore, we explore employing foundation models to facilitate self-supervised MDE. 
\vspace{-1mm}
\section{\ours}
We present \ours, a novel framework that first aggregates multi-grained features from CLIP (global semantics) and DINO (local spatial details) under contrastive language guidance (Sec.~\ref{sec: stage_1}).
Then, we further optimize the model with the help of the auxiliary camera pose information and align the hybrid-grained features with language guidance.
Additionally, it could serve as a plug-and-play depth encoder to improve the capture of continuous depth variations, thereby enhancing the accuracy of self-supervised MDE methods (Sec.~\ref{sec: stage_2}).

\vspace{-1mm}

\subsection{Coarse Depth Sensing}
\label{sec: stage_1}
As shown in Fig.~\ref{fig: method}, we use the geometric priors in self-driving scenes, where depth increases along lane markings, to serve as a practical proxy for depth supervision because lane labels are more accessible than precise depth measurements. \looseness=-1

\begin{figure}[H]
    \centering
    \includegraphics[width=1\linewidth]{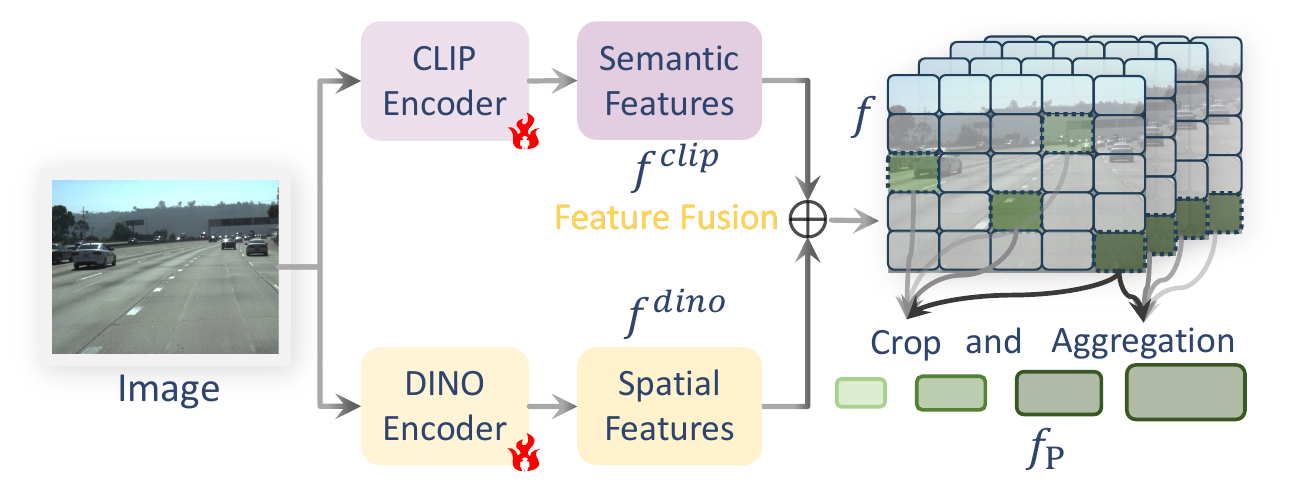}
    \vspace{-6mm}
    \caption{Patch selecting and feature concatenating.}
    \label{fig:ddl}
    \vspace{-4mm}
\end{figure}

Given an image I from the lane detection datasets such as TuSimple, we separately extract global semantics features $f^{\text{clip}}$ using CLIP and the spatial details features $f^{\text{dino}}$ with DINO, as illustrated in Fig.~\ref{fig:ddl}:
\begin{equation}
    \begin{aligned}
        f^{clip} &= \mathbf{F}^{clip}(\mathbf{I}) \\
        f^{dino} &= \mathbf{F}^{dino}(\mathbf{I}) .
    \label{eq:patch}
    \end{aligned}
\end{equation}

Considering the final features from the model's last layer have limited information, we obtain multi-scale feature maps from four ResNet blocks of CLIP and DINO 2-, 5-, 8-, and 11-th layers.
To match the spatial dimensions of the DINO features, we interpolate $f^{clip}$ to the size of $f^{dino}$ and then concatenate them to acquire the fused features $f$.
\vspace{-2mm}

\subsubsection{Patch Selecting}
We crop a set of patches $f_{\mathbf{P}} = [f_{\mathbf{P}_0}, f_{\mathbf{P}_1},..., f_{\mathbf{P}_{N-1}]}$ along the lane labels on the fused features $f$,  where $N$ is the number of patches (here, $N = 7$).
These selected patches adhere to the following rules:
\begin{itemize}
\item For any two patches ($\mathbf{P}_i$ and $\mathbf{P}_j$, $0 \leq i < j \leq N-1 $), the position of $\mathbf{P}_i$ must be higher (\eg, smaller y coordinate) than that of $\mathbf{P}_j$.
\item They are chosen randomly on the column and regularly internally across the row. \looseness=-2
This sample strategy could mitigate the issue of patches exhibiting nearly identical depth since the dataset is ego-view and vehicles at similar depths typically appear in the same row.

\end{itemize}


\subsubsection{Depth Prompt Design}
The original CLIP is not designed for depth estimation, it is necessary to customize effective depth text prompts.
As mentioned above, we have successfully
collected a series of patches whose depths are ordinal/ordered.
The depth of different patches is ordinal, and the higher patches correspond to a more or equal depth. 
Therefore, we construct ranking-depth text prompts to describe the ordinal relationship of image patches. 
The depth text prompt is formatted as ``This patch appears to be [\textit{depth token}]'', where [\textit{depth token}] represents a set of rank distances like ``\textit{very distant, ..., close, very close}''. 
These prompts are then fed into the text encoder to output depth text rank embeddings $\mathbf{T} = [\mathbf{T}_0, \ldots, \mathbf{T}_{N-1}]\in \mathbb{R}^{N \times C}$.

\subsubsection{Contrastive Learning}

Unlike the original CLIP~\cite{clip}, which relies solely on a single contrastive mechanism, 
We introduce two complementary types of contrastive learning loss to refine its ability to differentiate depth-related features:
\begin{itemize}
    \item \textbf{Intramodal Contrastive Learning}: 
    Within the visual modality, we enforce depth-aware representation learning by defining constraints on patch embeddings. 
    Given a set of extracted features $f_{\mathbf{P}}$, we use the patch index to guide the feature alignment. 
    Formally, we compute the similarity between patches as follows:
    \begin{equation}
    s^{intra}_{i,j} = f_{\mathbf{P}_i} \cdot f_{\mathbf{P}_j}' ,
    \end{equation}
    where $ 0 \leq i \leq N-1 $ and $ 0 \leq j \leq N-1 $. 
    We introduce an intramodal contrastive loss to ensure that the self-similarity of a patch (which serves as a baseline) is higher than its similarity with patches corresponding to different depth levels. Specifically, we hope the similarity matrix $\mathbf{S}$ is a specific ordinal matrix as shown in Fig.~\ref{fig: method}:
    \begin{equation}
    \mathcal{L}_i = \sum_{i=0}^{N-1} \max\Bigl(0,\, s^{intra}_{i',i} - s^{intra}_{i,i}\Bigr) ,
    \end{equation}
    where $ 0 \leq i^{'} \leq i \leq N-1$.
    This loss penalizes cases where a patch with a lower index (assumed to be closer) is less similar to its own representation than a patch with a higher index (assumed to be farther).
    It ensures that the learned features reflect the natural depth ordering present in the image.
    \item  \textbf{Language-guided Contrastive Learning}: 
    In addition to intramodal contrastive learning, we use language-guided contrastive learning to align visual features with corresponding depth-related text embeddings. In our framework, each patch is associated with a textual descriptor (\eg, ``near'' or ``distant'') based on its relative depth position, which is determined by the patch index, which serves as a proxy for depth. This association allows us to ensure that a patch's visual features are more similar to its corresponding text embedding than to other depth descriptors.
    Furthermore, the depth-related text can serve as a granularity calibrator to promote the fusion of CLIP and DINO.
    We first compute the similarity matrix $\mathbf{S} \in \mathbb{R}^{N \times N}$ with elements defined as:
    \begin{equation}
    s^{cross}_{i,j} = f_{\mathbf{P}_i} \cdot \mathbf{T}_j' .
    \end{equation}
    Due to the inheritance of ranking relationships from the patches and depth text prompts, we hope the similarity matrix $\mathbf{S}$ is a specific ordinal matrix as follows:
    \begin{equation}
    \begin{aligned}
    & s^{cross}_{i',i} \leq s^{cross}_{i,i} \\[2mm]
    \mathcal{L}_c &= \sum_{i=0}^{N-1} \max\Bigl(0,\, s^{cross}_{i',i} - s^{cross}_{i,i}\Bigr) ,
    \end{aligned}
    \end{equation}
\end{itemize}
where $0 \leq i^{'} \leq i \leq N-1 $.
This cross-modal contrastive loss ensures that the visual features are well aligned with the semantic depth cues provided by the text, further reinforcing depth awareness.\looseness=-1
    
By integrating both intramodal and language-guided contrastive learning, we enable the CLIP and DINO encoders to effectively capture and distinguish depth variations. 
Notably, only the visual encoder is trained, while the text encoder remains frozen during this stage.

\subsection{Fine Depth Estimation}
\label{sec: stage_2}
In this phrase, we obtain a dense depth map by aligning the aggregated features with learnable depth instruction and combining an auxiliary camera pose from multi-frames as shown in Fig.~\ref{fig: method}(b).
Classically, previous methods \cite{zhou2017,monodepthv2, bian2021unsupervised, wang2023planedepth} generally formulate self-supervised MDE as the minimization of a photometric reprojection error during training.
For the random two consecutive frames $(I_a, I_b)$ from the training video, they utilize separate branches\;-\;DepthNet and PoseNet\;-\;to extract depth feature and pose feature.
Based on these two features, they predict the depth map $\hat{D}_a$ for $I_a$ and estimate their relative 6-DoF camera pose $M_{a\rightarrow b}$.

\subsubsection{Learnable Depth Prompt}
To mitigate the limitations of human language \cite{COOP, auty2023learning}, we replace manually crafted depth tokens like ``\textit{close, far, ...}'' with $N$ learnable tokens. 
The learnable depth tokens are initialized by randomly sampling 512 elements from a normal distribution with a mean of 0 and a standard deviation of 0.02.
After being processed by the text encoder, we obtain the depth text embedding $\mathbf{g} \in \mathbb{R}^{  N \times C }$.

\subsubsection{Pixel-level Language Alignment}
Given input $\mathbf{I}$ from the self-supervised MDE dataset, we first aggregate hybrid-grained features from DINO and CLIP to acquire four scales aggregated feature $\mathbf{f}^d$.
To maintain the global semantic understanding of CLIP and the local spatial reasoning of DINO, we employ depth instruction as a granularity calibrator by aligning the aggregated features $\mathbf{f}$ with the corresponding learnable depth prompt:
\begin{equation}
\mathbf{f}^d = \operatorname{ALIGN}\left\langle \mathbf{f}, \mathbf{g} \right\rangle .
\end{equation}
Specially, for the operation $\operatorname{ALIGN}\left\langle \mathbf{f}, \mathbf{g} \right\rangle $, we adopt following steps: 
\begin{itemize}
    \item Reshape $\mathbf{f}$: Initially, we reshape the $\mathbf{f}$ into a matrix with dimensions $C \times HW $.
    \item Inner Product with Depth Text Embedding: we compute the inner product between $\mathbf{f} $ and the transpose of the depth text embedding. This operation yields a new matrix ${\mathbf{f}}$ with dimensions $N\times HW$.
    \item Recover Reshaping: We reshape $\mathbf{f}$ back into the desired feature tensor $\mathbf{f}^d$ and replace the depth feature from the original DepthNet in existing self-supervised MDE methods.\looseness=-1
\end{itemize}

Finally, we predict depth map $\hat{D}$ by up-sampling $\mathbf{f}$ following DPT~\cite{dpt21} with the help of auxiliary camera pose.
\begin{table*}[h]
 \resizebox{\textwidth}{!}{%
    \scriptsize
    \begin{tabular}{ l | c | c | c | c | c | c | c | c | c }
        \toprule
         Method & W $\times$ H & Train & Abs Rel  $\downarrow$ &  Sq Rel   $\downarrow$ & RMSE  $\downarrow$ &  \begin{tabular}[c]{@{}c@{}}RMSE \\ log \end{tabular} $\downarrow$ &  $\delta$ \textless 1.25 $\uparrow$  &  $\delta$ \textless $1.25^2$   $\uparrow$ &  $\delta$ \textless $1.25^3$ $\uparrow$ \\
        \midrule
        GeoNet \cite{yin2018geonet} & $ 640 \times 192 $ & M & 0.149 &1.060 &5.567 &0.226& 0.796& 0.935 &0.975 \\

        Johnston \textit{et al.} \cite{Johnston2020CVPR} & $ 640 \times 192 $ & M & 0.106 & 0.861 & 4.699 & 0.185 & 0.889 & 0.962 & 0.982 \\
        PackNet-SfM \cite{packnet} & $640 \times 192$ & M & 0.111 & {0.785} & {4.601} & {0.189} & 0.878 & {0.960} & {0.982} \\

        MonoFormer \cite{bae2022monoformer} &  $640\times192$ &M & 0.104 & 0.846 & 4.580 & 0.183 & 0.891 & 0.962 &0.982 \\
        
        DIFFNET \cite{DIFF21} & $640 \times 192$ & M & 0.102 & {0.764} & {4.483} & {0.180} & 0.896 & {0.965} & {0.983} \\

        BRNet \cite{brnet22} & $640 \times 192$ & M & 0.105 & {0.698} & {4.462} & {0.179} & 0.890 & {0.965} & 0.984 \\
        
        FeatureNet \cite{shu2020feature} & $640\times192$ & M &  0.104 & 0.729 & 4.481 & 0.179 & 0.893 & 0.965 & 0.984 \\
        
        MonoViT  \cite{zhao2022monovit} & $640\times192$ & M & 0.099 & {0.708} & {4.372} & {0.175} & 0.900 & {0.967} & 0.984 \\ 

        Lite-Mono \cite{litemono23} & $640\times192$ & M & 0.107 & {0.765} & {4.561} & {0.183} & 0.886 & {0.963} & {0.983}
        \\
        DualRefine \cite{dualpathdepth25} & $640\times192$ & M & 0.103 & 0.776 & 4.491 & 0.181 & 0.894 & 0.965 & 0.983  \\
        Dynamicdepth \cite{dyamicdepth22} & $640\times192$ & M & 0.096 & {0.720} & {4.458} & {0.175} & 0.897 & {0.964} & 0.984 \\
        SQLdepth \cite{WangLiangXuJiaoYu2024} & $640\times192$ & M & \underline{0.094} & {0.697} & {4.320} & {0.172} & {0.904} & {0.967} & 0.984 \\ 

       RPrDepth \cite{han2025high} & $640\times192$ & M & 0.097 & {0.658} & {4.279} & \underline{0.169} & 0.900 & {0.967} & \textbf{0.985} \\
       
        \midrule

        Manydepth \cite{manydepth21} & $640\times192$ & M & 0.098 & {0.770} & {4.459} & {0.176} & 0.900 & 0.965 & 0.983 \\
        
       \rowcolor{linecolor2}\quad w/ \ours & $640\times192$ & M & 0.096 & {0.665} & {4.192} & 0.170 & \underline{0.906} & \underline{0.968} & \textbf{0.985}  \\

       Mono-ViFI \cite{liu2024mono} & $640\times192$ & M & 0.105 & 0.708 & 4.446 & 0.179 & 0.887 & 0.965 & 0.984 \\

       \rowcolor{linecolor2}\quad w/ \ours & $640\times192$ & M & \underline{0.094} & \underline{0.658} & \underline{4.168} & \underline{0.169} & \underline{0.906} & \underline{0.968} & \textbf{0.985} \\

        Monodepth2 \cite{monodepthv2} & $ 640 \times 192 $ & M  & 0.115 & 0.903 & 4.863 & 0.193 & 0.877 & 0.959 & 0.981 \\
        \rowcolor{linecolor2}\quad w/  \ours  & $640\times192$ & M & \textbf{0.093}  &  \textbf{0.596}  &  \textbf{4.113}  &  \textbf{ 0.167}  &   \textbf{0.910}  &   \textbf{0.970}  &   \textbf{0.986}  \\

        \bottomrule
    \end{tabular}
    }
    \centering
    \vspace{-3mm}
    \caption{\textbf{Quantitative results} on  KITTI Eigen split \cite{eigen2014depth} with the state-of-the-art self-supervised MDE methods. ``$W \times H$'' denotes the resolution of input images. The column of ``train'' specifies the way of training, where ``M'' represents optimizing the model using monocular video. All results are Post-Processed~\cite{godard2017unsupervised}. The best results are in \textbf{bold}; the second best is \underline{underlined}. The methods integrate with \ours~modules and outperform all previous methods by a large margin on all metrics.}
    \label{tab:kitti_res}
\end{table*}

\begin{table*}[h]
 \resizebox{\textwidth}{!}{%
    \scriptsize
    \begin{tabular}{ l | c | c | c | c | c | c | c | c | c }
        \toprule
         Method & W $\times$ H & Train & Abs Rel  $\downarrow$ &  Sq Rel   $\downarrow$ & RMSE  $\downarrow$ &  \begin{tabular}[c]{@{}c@{}}RMSE \\ log \end{tabular} $\downarrow$ &  $\delta$ \textless 1.25 $\uparrow$  &  $\delta$ \textless $1.25^2$   $\uparrow$ &  $\delta$ \textless $1.25^3$ $\uparrow$ \\
        \midrule
        DepthCLIP \cite{depthclip} & - & 0-shot & 0.473 & 6.007 & 12.958 &0.680 & 0.281 & 0.531 & 0.696  \\
        Hu \textit{et al.}~\cite{hu2024learning} &  $704 \times 352 $ & 1-shot & 0.384 & 4.661 & 12.290 & 0.632 & 0.312 & 0.569 & 0.739 \\
        Auty \textit{et al.} \cite{auty2023learning} & $640 \times 192$ & M & 0.303 & 6.322 & - & - & 0.550 & 0.830 & 0.938 \\ 
       \rowcolor{linecolor2} Monodepth2 w/  \ours  & $640\times192$ & M & \textbf{0.093}  &  \textbf{0.596}  &  \textbf{4.113}  &  \textbf{ 0.167}  &   \textbf{0.910}  &   \textbf{0.970}  &   \textbf{0.986}  \\
        \bottomrule
    \end{tabular}
    }
    \centering
    \vspace{-3mm}
    \caption{Comparison with previous MDE methods using CLIP. We outperform all previous methods by a large margin on all metrics.}
    \vspace{-5mm}
    \label{tab:kitti_clip_res}
\end{table*}

\subsubsection{Reconstruction and Smoothness Loss}
Following previous methods \cite{zhou2017,monodepthv2}, 
we optimize the model using image reconstruction loss \cite{zhao2016loss} and smoothness loss~\cite{godard2017unsupervised}.
Concretely, we warp $I_b$ into $I_a$ to generate the reconstructed image $\hat{I_b}$ according to the predicted depth map $\hat{D}_a$ and camera pose $M_{a\rightarrow b}$ as following equation:
\begin{equation}
\begin{aligned}
\hat{I}_{b} & =I_{a}\left\langle\operatorname{proj}\left(\hat{D}_a, M_{a \rightarrow b}, K\right)\right\rangle ,
\end{aligned}
\end{equation}
where $\left\langle \right\rangle$ is the differentiable bilinear sampling operator following~\cite{monodepthv2}.
To evaluate the reconstructed images $\hat{I}_b$, we use L1-norm distance in pixel space and SSIM \cite{wang2004image} to construct photometric error function following \cite{zhao2016loss, godard2017unsupervised}:
\begin{equation}
\mathcal{L}_{p e}=\frac{\beta}{2}\left(1-\operatorname{SSIM}\left(I_b, \hat{I}_b\right)\right)+(1-\beta)\left\|I_b-\hat{I}_b\right\|_1,
\end{equation}
where $\beta= 0.85$ by default and SSIM computes pixel similarity over a $3\times3$ window.
To improve the smoothness of the predicted depth map, we adopt the edge-aware smoothness loss \cite{godard2017unsupervised} to regularize the predicted depth map:
\begin{equation}
\mathcal{L}_{smooth}= \left|\partial_x d_t^*\right|e^{-\left|\partial_x I_t\right|}+\left| \partial_y d_t^* \right| e^{-\left|\partial_y I_t\right|}.
\end{equation}
Overall, the total loss $\mathcal{L}_{total}$ for training is formulated as:
\begin{equation}
\mathcal{L}_{total} = \mathcal{L}_{pe} + \lambda*\mathcal{L}_{smooth},
\end{equation}
where $\lambda$ is set to 0.001 by default.
To reduce the trainable parameters, we only train the visual encoder, learnable tokens, and up-sample layers while freezing the text encoder.

\section{Experiments}

\vspace{-2mm}
\subsection{Implementation Details}
Unless otherwise specified, we use the ResNet-50-based CLIP \cite{clip} visual encoder and DINOVv2 \cite{dinov223} encoder as the backbone in the image branch and vanilla CLIP text encoder.
Our models are implemented in PyTorch \cite{pytorch} and trained for 10 epochs with a batch size of 16.
We adopt AdamW optimizer with an initial learning rate of 1e-4 and StepLR policy.
All the experiments are conducted on a single NVIDIA A800 GPU.
Moreover, the number of learnable depth and pose tokens is set to 256.

\begin{figure*}[tb]
  \centering
  \includegraphics[width=1.0\linewidth]{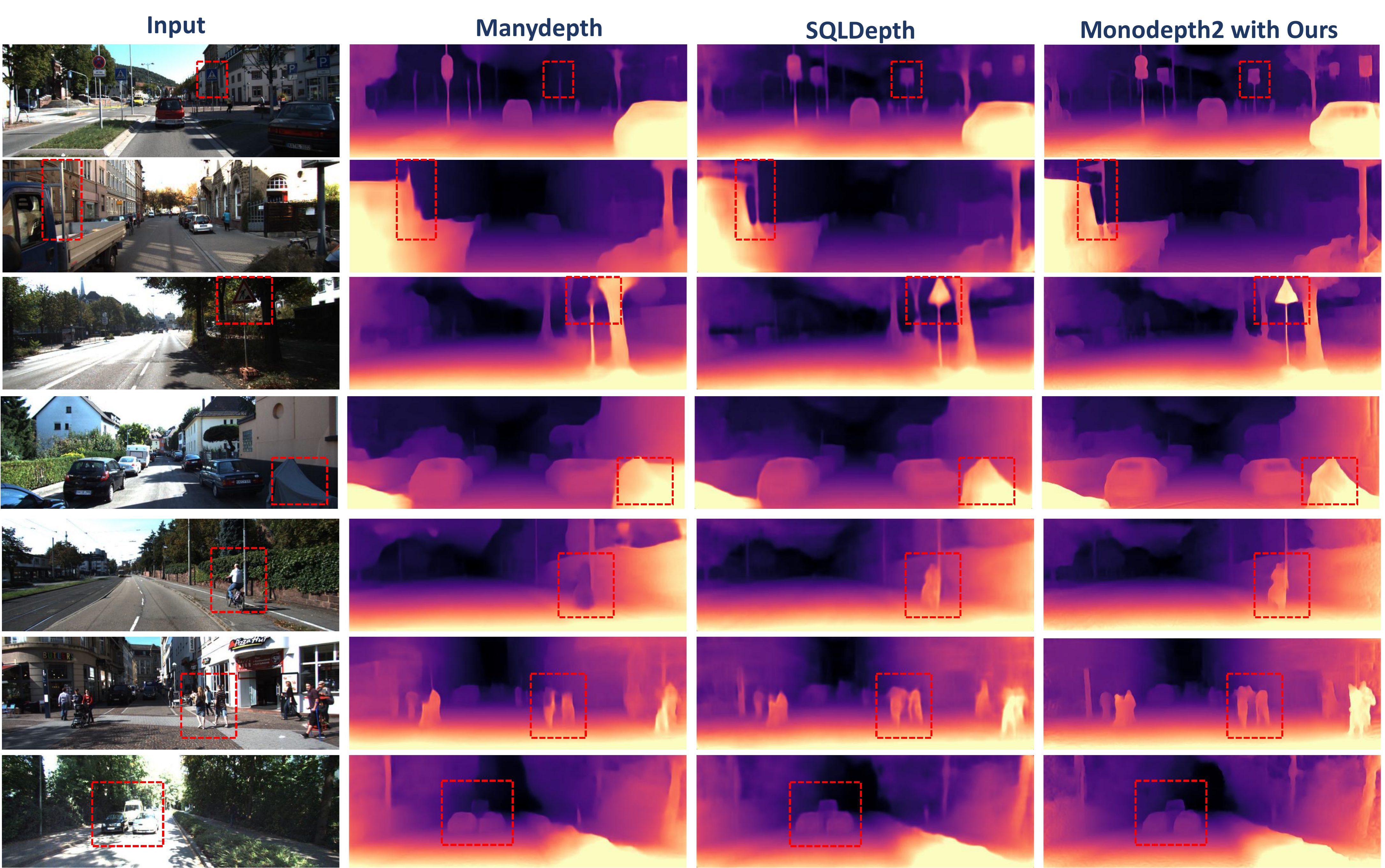}
  \vspace{-7mm}
  \caption{Qualitative comparison with Manydepth~\cite{manydepth21}, SQLDepth~\cite{WangLiangXuJiaoYu2024} and Monodepth2 with Hybrid-depth on the KITTI dataset. Monodepth2 with \ours~accurately predicts continuous depth in the ground region while preserving sharp object edges.}
  \label{fig: Q_result}
  \vspace{-6mm}
\end{figure*}

\subsection{Comparison with State-of-the-art}
\subsubsection{Results on KITTI}
Table~\ref{tab:kitti_res} presents a comparison between three classical MDE methods\;-\;Monodepth2~\cite{monodepthv2}, ManyDepth~\cite{manydepth21} and Mono\;-\;VIFI~\cite{liu2024mono}—augmented with our framework, and recent state-of-the-art self-supervised MDE approaches.
Our proposed method not only enhances the performance of the original methods but also outperforms all other approaches by a significant margin across all evaluation metrics.
In addition, Table~\ref{tab:kitti_clip_res} reports the same metrics for few-shot or supervised monocular depth estimation methods that utilize CLIP.
Although Auty \textit{et al.} \cite{auty2023learning} and Hu \textit{et al.} \cite{hu2024learning} achieve MDE in a supervised manner, their performance remains substantially lower than that of current self-supervised approaches.
In contrast, \ours~consistently achieves higher accuracy across all metrics, demonstrating that our method effectively leverages the power of large-scale models and pre-training with depth contrastive learning.
Additionally, we find \ours~promotes the convergence of the existing self-supervised MDE methods.

\subsubsection{Qualitative Results}
Fig.~\ref{fig: Q_result} presents qualitative comparisons between Monodepth2 with \ours~and previous approaches on challenging KITTI images. 
The results demonstrate that Monodepth2 with \ours~ achieves superior structural and edge preservation compared to its counterparts. 
It produces smoother depth estimations in ground regions while maintaining sharp object boundaries, effectively recovering fine details, and significantly reducing depth artifacts around object edges.

\subsection{Ablation Studies}
In this section, we design the experiments to investigate the following questions:


\question \textbf{Do gains stem from the coarse depth sensing stage, not backbone improvements?}
To investigate the source of performance improvement, we conduct experiments as illustrated in Table.~\ref{tab:ab_dc}. 
We compare the results of our approach with those obtained using the same backbone but without the coarse depth sensing stage. 
The findings reveal that merely using a more powerful backbone does not achieve the same level of enhancement. 
In contrast, the coarse-to-fine learning scheme boosts performance. 
\vspace{-4mm}
\begin{table}[h]
\resizebox{0.48\textwidth}{!}{%
    \scriptsize
    \setlength{\tabcolsep}{2pt}
    \begin{tabular}{ l | c | c | c | c | c }
        \toprule
         Method & Abs Rel  $\downarrow$ &  Sq Rel   $\downarrow$ & RMSE  $\downarrow$ &  \begin{tabular}[c]{@{}c@{}}RMSE \\ log \end{tabular} $\downarrow$ &  $\delta$ \textless 1.25 $\uparrow$   \\
        \midrule
         Manydepth \cite{manydepth21} & 0.098 & {0.770} & {4.459} & {0.176} & 0.900 \\
         \quad w/o co & 0.096 & 0.725 & 4.374 & 0.173 & 0.905 \\
         \rowcolor{linecolor2} \quad w/ \ours & 0.096 & 0.665 & 4.192 & 0.170 & 0.906 \\
         \midrule
         Monodepth2 \cite{monodepthv2} & 0.115 & 0.903 & 4.863 & 0.193 & 0.877 \\
        \quad w/o co & 0.105 & 0.752 & 4.455 & 0.182 & 0.902 \\
         \rowcolor{linecolor2} \quad w/ \ours & \textbf{0.093}  &  \textbf{0.596}  &  \textbf{4.113}  &  \textbf{ 0.167}  &   \textbf{0.910}   \\
        \bottomrule
    \end{tabular}
    }
    \centering
    \vspace{-3mm}
    \caption{\looseness=-1 Impact of coarse depth sensing stage versus backbone strength on performance, where ``w/o co'' denotes that directly train the models following existing self-supervised MDE methods.}
    \vspace{-4mm}
    \label{tab:ab_dc}
\end{table}

\question \textbf{Is it necessary for language-guided contrastive learning?}
To evaluate the effectiveness of language-guided contrastive, we conducted ablation studies by removing the cross-modal contrastive loss to compare the performance against the full model.
Furthermore, we explored the role of intramodal contrastive learning.
As shown in Table~\ref{tab:ab_loss}, the model without feature alignment exhibited a noticeable decline in performance across all metrics. 
This suggests that aligning visual features with depth-related text cues and intramodal contrastive learning effectively enhances the model's understanding of spatial relationships and improves depth prediction.

\begin{table}[H]
\resizebox{0.48\textwidth}{!}{%
    \scriptsize
    \setlength{\tabcolsep}{2pt}
    \begin{tabular}{ l | c | c | c | c | c }
        \toprule
         Method & Abs Rel  $\downarrow$ &  Sq Rel   $\downarrow$ & RMSE  $\downarrow$ &  \begin{tabular}[c]{@{}c@{}}RMSE \\ log \end{tabular} $\downarrow$ &  $\delta$ \textless 1.25 $\uparrow$   \\
        \midrule
        ManyDepth \cite{manydepth21} & 0.098 & {0.770} & {4.459} & {0.176} & 0.900 \\
        \quad w/ $L_i$  & 0.098  &   0.717 &   4.327  &   0.173  &   0.905  \\ 
        \quad w/ $L_c$ & 0.096 & 0.667 & 4.199 & 0.174 & 0.905 \\
        \rowcolor{linecolor2} \quad w/ \ours & 0.096 & 0.665 & 4.192 & 0.170 & 0.906 \\
        \midrule
        
        Monodepth2 \cite{monodepthv2} & 0.115 & 0.903 & 4.863 & 0.193 & 0.877 \\

        \quad w/ $L_i$  & 0.098 & 0.663 & 4.236 & 0.172 & 0.902 \\ 
        
        \quad w/ $L_c$ & 0.095 & 0.675 & 4.160 & 0.169 & 0.909 \\ 

        \rowcolor{linecolor2} \quad w/ \ours & \textbf{0.093}  &  \textbf{0.596}  &  \textbf{4.113}  &  \textbf{ 0.167}  &   \textbf{0.910}    \\
        \bottomrule
    \end{tabular}
    }
    \vspace{-3mm}
    \centering
    \caption{Comparison of the results with different contrastive loss.}
    \vspace{-4mm}
    \label{tab:ab_loss}
\end{table}

\question \textbf{Is depth instruction necessary as a granularity calibrator?}
To assess the role of depth instruction as a granularity calibrator, we conducted an ablation study by removing the depth instruction component in both stages. In the first stage, we disabled the cross-modal contrastive loss aligning CLIP-DINO features with depth-related text cues, while the fusion process and intramodal contrastive learning remained unchanged. 
During the fine depth estimation phrase, we used a naive fusion method without language alignment for CLIP-DINO features. 
The results in Table~\ref{tab:ab_gc} show a significant performance drop when depth instruction is removed, confirming its importance for calibrating granularity between CLIP and DINO features.
\vspace{-4mm}
\begin{table}[H]
\resizebox{0.48\textwidth}{!}{%
    \scriptsize
    \setlength{\tabcolsep}{2pt}
    \begin{tabular}{ l | c | c | c | c | c }
        \toprule
         Method & Abs Rel  $\downarrow$ &  Sq Rel   $\downarrow$ & RMSE  $\downarrow$ &  \begin{tabular}[c]{@{}c@{}}RMSE \\ log \end{tabular} $\downarrow$ &  $\delta$ \textless 1.25 $\uparrow$   \\
        \midrule
        Monodepth2 \cite{monodepthv2} & 0.115 & 0.903 & 4.863 & 0.193 & 0.877 \\

        \quad w/o co-gc & 0.098 & 0.663 & 4.236 & 0.172 & 0.904 \\ 
        
        \quad w/o fi-gc &  0.100  &   0.717  &   4.284  &   0.173  &   0.905 \\ 

        \rowcolor{linecolor2} \quad w/ \ours & \textbf{0.093}  &  \textbf{0.596}  &  \textbf{4.113}  &  \textbf{ 0.167}  &   \textbf{0.910}    \\
        \bottomrule
    \end{tabular}
    }
    \vspace{-2mm}
    \centering
    \caption{Comparison of the results when using depth instruction as a granularity calibrator in both stages, where ``w/o co-gc'' and ``w/o fi-gc'' represent feature fusion without alignment during coarse and fine stage, respectively.}
    \label{tab:ab_gc}
\end{table}  
\vspace{-4mm}

\question \textbf{Why do we need both CLIP and DINO?}
To address this question, we conducted ablation experiments comparing models that use only CLIP-RN50 or DINOv2 VIT-B encoder versus our dual-encoder approach as shown in Table.~\ref{tab:ab_clipdino}.  
Our results, combined with Monodepth2~\cite{monodepthv2}, reveal that relying solely on a single encoder significantly degrades performance. 
CLIP provides robust semantic understanding, whereas DINO excels at extracting fine-grained spatial details. 
By integrating both encoders, our model leverages complementary strengths to facilitate depth estimation.
\vspace{-3mm}
\begin{table}[H]
\resizebox{0.48\textwidth}{!}{%
    \scriptsize
    \setlength{\tabcolsep}{2pt}
    \begin{tabular}{ l | c | c | c | c | c }
        \toprule
         Method & Abs Rel  $\downarrow$ &  Sq Rel   $\downarrow$ & RMSE  $\downarrow$ &  \begin{tabular}[c]{@{}c@{}}RMSE \\ log \end{tabular} $\downarrow$ &  $\delta$ \textless 1.25 $\uparrow$   \\
        \midrule
        Monodepth2 \cite{monodepthv2} & 0.115 & 0.903 & 4.863 & 0.193 & 0.877  \\
        \quad w/ CLIP  & 0.102 & 0.752 & 4.667 & 0.182 & 0.902 \\
        \quad w/ DINO   & 0.104 & 0.769 & 4.685 & 0.180 & 0.903  \\ 
        \rowcolor{linecolor2} \quad w/ \ours~ & \textbf{0.093}  &  \textbf{0.596}  &  \textbf{4.113}  &  \textbf{ 0.167}  &   \textbf{0.910}    \\
        \bottomrule
    \end{tabular}
    }
    \vspace{-3mm}
    \centering
    \caption{Comparison of the result with employing CLIP, DINO, and CLIP+DINO.}
    \vspace{-4mm}
    \label{tab:ab_clipdino}
\end{table}

\question \textbf{What is the effect of learnable depth tokens' count?}
As shown in Table~\ref{tab:ab_tokennum}, we investigate the impact of the number of learnable depth tokens by varying the token count in Monodepth2~\cite{monodepthv2} integrated with \ours. 
The results demonstrate a non-monotonic relationship: performance improves with increasing token count but degrades beyond this optimal range (e.g., 256 tokens). This trend arises due to a capacity-overfitting trade-off: while more tokens enhance model flexibility, excessive tokens may over-parameterize the depth prior space, causing overfitting. Despite this, \ours~exhibits strong robustness and outperforms previous self-supervised methods, thanks to our depth-aware prompt calibration mechanism.
\vspace{-3mm}
\begin{table}[H]
\resizebox{0.48\textwidth}{!}{%
    \scriptsize
    \setlength{\tabcolsep}{3.5pt}
    \begin{tabular}{ l | c | c | c | c | c }
        \toprule
         Method & Abs Rel  $\downarrow$ &  Sq Rel   $\downarrow$ & RMSE  $\downarrow$ &  \begin{tabular}[c]{@{}c@{}}RMSE \\ log \end{tabular} $\downarrow$ &  $\delta$ \textless 1.25 $\uparrow$   \\
        \midrule
        128 & 0.095  &   0.650  &   4.199  &   0.170  &   \textbf{0.911}  \\
        256 & \textbf{0.093}  &  \textbf{0.596}  &  \textbf{4.113}  &  \textbf{ 0.167}  &   0.910 \\ 
        512 & 0.097  &   0.658  &   4.182  &   0.172  &   0.908  \\
        1024 & 0.095  &   0.649  &   4.154  &   0.170  & \textbf{0.911}  \\
        \bottomrule
    \end{tabular}
    }
    \vspace{-3mm}
    \centering
    \caption{Comparison of the results with variant token counts.}
    \label{tab:ab_tokennum}
\end{table}    
\vspace{-4mm}

\question \textbf{Can \ours~benefit other 3D perception tasks such as BEV prediction?}
\label{sec:bev}
We integrate \ours~with established BEV perception methods like BEVDet \cite{huang2021bevdet} and FB-BEV~\cite{li2023fbbev}. 
For fair comparison, we report results using the simplest baseline configurations for both BEVDet and FB-BEV. As shown in Table~\ref{tab:bev}, our approach significantly improves the performance of both methods. These findings reveal that \ours~could effectively promote 3D perception.\looseness=-1
\vspace{-3mm}
\begin{table}[H]
\centering
\resizebox{0.46\textwidth}{!}{%
    \scriptsize
    \setlength{\tabcolsep}{4pt}
    \begin{tabular}{l|c|ccc|c}
    \toprule
    Method    & mAP$\uparrow$  & mATE$\downarrow$ & mASE$\downarrow$  & mAOE$\downarrow$ &  NDS $\uparrow$ \\
    \midrule
    BEVDet \cite{huang2021bevdet}                             & 0.283 & 0.773 & 0.288 & 0.698 & 0.350 \\
    \rowcolor{linecolor2} \quad w/ \ours  & \textbf{0.325} & \textbf{0.734} & \textbf{0.281} & \textbf{0.623}  & \textbf{0.395} \\
    \midrule
    FB-BEV \cite{li2023fbbev}                            & 0.312 &  0.702 & 0.275 & 0.518  & 0.406 \\
    \rowcolor{linecolor2} \quad w/ \ours                        & \textbf{0.348} & \textbf{0.673} &\textbf{ 0.259} & \textbf{0.464}  & \textbf{0.439} \\
    \bottomrule
    \end{tabular}
}
\vspace{-2mm}
\caption{Comparison on the nuScenes \cite{caesar2020nuscenes} \emph{val} set. The resolution of the input image is set to $256\times704$. }
\label{tab:bev}
\end{table}
\vspace{-4mm}

\section{Conclusions}
In this paper, we present a novel paradigm named \ours~ for self-supervised MDE, which progressively equips the model with the ability to estimate depth from coarse to fine.
Initially, we aggregate different-grained features from CLIP and DINO for coarse depth sensing under contrastive language guidance.
Then, we fine-tune the model with an auxiliary camera pose to conduct fine depth estimation, and it could serve as an efficient depth encoder to improve the capture of continuous depth variations, thereby enhancing the accuracy of self-supervised MDE methods.
Additionally, we employ depth instruction as a granularity calibrator mechanism in both stages to address the issue of granularity mismatch to help hybrid-grained features to be well harmonized together.
This paradigm significantly outperforms all existing methods by a large margin while facilitating 3D recognition like BEV perception.\looseness=-1

\section*{Acknowledgements}
This work was supported by Grants of NSFC 62302246, ZJNSFC LQ23F010008, Ningbo 2023Z237 \& 2024Z284 \& 2024Z289 \& 2023CX050011 \& 2025Z038, and supported by High Performance Computing Center at Eastern Institute of Technology and Ningbo Institute of Digital Twin. 

{
    \small
    \bibliographystyle{ieeenat_fullname}
    \bibliography{main.bib}

\begin{thebibliography}{147}
\providecommand{\natexlab}[1]{#1}
\providecommand{\url}[1]{\texttt{#1}}
\expandafter\ifx\csname urlstyle\endcsname\relax
  \providecommand{\doi}[1]{doi: #1}\else
  \providecommand{\doi}{doi: \begingroup \urlstyle{rm}\Url}\fi

\bibitem[Auty and Mikolajczyk(2023)]{auty2023learning}
Dylan Auty and Krystian Mikolajczyk.
\newblock Learning to prompt clip for monocular depth estimation: Exploring the limits of human language.
\newblock In \emph{Proceedings of the IEEE/CVF International Conference on Computer Vision}, pages 2039--2047, 2023.

\bibitem[Bae et~al.(2023)Bae, Moon, and Im]{bae2022monoformer}
Jinwoo Bae, Sungho Moon, and Sunghoon Im.
\newblock Deep digging into the generalization of self-supervised monocular depth estimation.
\newblock In \emph{Proceedings of the AAAI Conference on Artificial Intelligence}, 2023.

\bibitem[Bhat et~al.(2021)Bhat, Alhashim, and Wonka]{adabins}
Shariq~Farooq Bhat, Ibraheem Alhashim, and Peter Wonka.
\newblock Adabins: Depth estimation using adaptive bins.
\newblock In \emph{CVPR}, 2021.

\bibitem[Bhat et~al.(2023)Bhat, Birkl, Wofk, Wonka, and M{\"u}ller]{bhat2023zoedepth}
Shariq~Farooq Bhat, Reiner Birkl, Diana Wofk, Peter Wonka, and Matthias M{\"u}ller.
\newblock Zoedepth: Zero-shot transfer by combining relative and metric depth.
\newblock \emph{arXiv preprint arXiv:2302.12288}, 2023.

\bibitem[Bian et~al.(2021)Bian, Zhan, Wang, Li, Zhang, Shen, Cheng, and Reid]{bian2021unsupervised}
Jia-Wang Bian, Huangying Zhan, Naiyan Wang, Zhichao Li, Le Zhang, Chunhua Shen, Ming-Ming Cheng, and Ian Reid.
\newblock Unsupervised scale-consistent depth learning from video.
\newblock \emph{International Journal of Computer Vision}, 129\penalty0 (9):\penalty0 2548--2564, 2021.

\bibitem[Bochkovskii et~al.(2024)Bochkovskii, Delaunoy, Germain, Santos, Zhou, Richter, and Koltun]{depthpro24}
Aleksei Bochkovskii, Ama\"{e}l Delaunoy, Hugo Germain, Marcel Santos, Yichao Zhou, Stephan~R. Richter, and Vladlen Koltun.
\newblock Depth pro: Sharp monocular metric depth in less than a second.
\newblock \emph{arXiv}, 2024.

\bibitem[Caesar et~al.(2020)Caesar, Bankiti, Lang, Vora, Liong, Xu, Krishnan, Pan, Baldan, and Beijbom]{caesar2020nuscenes}
Holger Caesar, Varun Bankiti, Alex~H Lang, Sourabh Vora, Venice~Erin Liong, Qiang Xu, Anush Krishnan, Yu Pan, Giancarlo Baldan, and Oscar Beijbom.
\newblock nuscenes: A multimodal dataset for autonomous driving.
\newblock In \emph{Proceedings of the IEEE/CVF conference on computer vision and pattern recognition}, pages 11621--11631, 2020.

\bibitem[Caron et~al.(2021)Caron, Touvron, Misra, J{\'e}gou, Mairal, Bojanowski, and Joulin]{dinov121}
Mathilde Caron, Hugo Touvron, Ishan Misra, Herv{\'e} J{\'e}gou, Julien Mairal, Piotr Bojanowski, and Armand Joulin.
\newblock Emerging properties in self-supervised vision transformers.
\newblock In \emph{Proceedings of the IEEE/CVF international conference on computer vision}, pages 9650--9660, 2021.

\bibitem[Casser et~al.(2019)Casser, Pirk, Mahjourian, and Angelova]{struct2depth}
Vincent Casser, Soeren Pirk, Reza Mahjourian, and Anelia Angelova.
\newblock Depth prediction without the sensors: Leveraging structure for unsupervised learning from monocular videos.
\newblock In \emph{Proceedings of the AAAI conference on artificial intelligence}, pages 8001--8008, 2019.

\bibitem[Chawla et~al.(2024)Chawla, Varma, Arani, and Zonooz]{Chawla2024WACV}
Hemang Chawla, Arnav Varma, Elahe Arani, and Bahram Zonooz.
\newblock Continual learning of unsupervised monocular depth from videos.
\newblock In \emph{Proceedings of the IEEE/CVF Winter Conference on Applications of Computer Vision (WACV)}, pages 8419--8429, 2024.

\bibitem[Chen et~al.(2022)Chen, Yao, Song, Li, Rao, and Zhang]{chen2022prompt}
Guangyi Chen, Weiran Yao, Xiangchen Song, Xinyue Li, Yongming Rao, and Kun Zhang.
\newblock Prompt learning with optimal transport for vision-language models.
\newblock \emph{arXiv preprint arXiv:2210.01253}, 2022.

\bibitem[Chen et~al.(2024)Chen, Liu, Li, Zhu, Wang, and Wu]{d3depth}
Siyu Chen, Hong Liu, Wenhao Li, Ying Zhu, Guoquan Wang, and Jianbing Wu.
\newblock D3epth: Self-supervised depth estimation with dynamic mask in dynamic scenes.
\newblock \emph{arXiv preprint arXiv:2411.04826}, 2024.

\bibitem[Cheng et~al.(2024)Cheng, Han, Liang, Wang, Zhang, and Liu]{cheng2024self}
Zhiyuan Cheng, Cheng Han, James Liang, Qifan Wang, Xiangyu Zhang, and Dongfang Liu.
\newblock Self-supervised adversarial training of monocular depth estimation against physical-world attacks.
\newblock \emph{arXiv preprint arXiv:2406.05857}, 2024.

\bibitem[Dong et~al.(2023{\natexlab{a}})Dong, Qi, Zhang, Zhang, Sun, Ge, Yi, and Ma]{act23}
Runpei Dong, Zekun Qi, Linfeng Zhang, Junbo Zhang, Jianjian Sun, Zheng Ge, Li Yi, and Kaisheng Ma.
\newblock Autoencoders as cross-modal teachers: Can pretrained 2d image transformers help 3d representation learning?
\newblock In \emph{The Eleventh International Conference on Learning Representations (ICLR)}, 2023{\natexlab{a}}.

\bibitem[Dong et~al.(2023{\natexlab{b}})Dong, Bao, Zheng, Zhang, Chen, Yang, Zeng, Zhang, Yuan, Chen, et~al.]{dong2023maskclip}
Xiaoyi Dong, Jianmin Bao, Yinglin Zheng, Ting Zhang, Dongdong Chen, Hao Yang, Ming Zeng, Weiming Zhang, Lu Yuan, Dong Chen, et~al.
\newblock Maskclip: Masked self-distillation advances contrastive language-image pretraining.
\newblock In \emph{Proceedings of the IEEE/CVF Conference on Computer Vision and Pattern Recognition}, pages 10995--11005, 2023{\natexlab{b}}.

\bibitem[Dong et~al.(2024{\natexlab{a}})Dong, Guo, Liu, Zhang, and Zhang]{dong2024ppea}
Yue-Jiang Dong, Yuan-Chen Guo, Ying-Tian Liu, Fang-Lue Zhang, and Song-Hai Zhang.
\newblock Ppea-depth: Progressive parameter-efficient adaptation for self-supervised monocular depth estimation.
\newblock In \emph{Proceedings of the AAAI Conference on Artificial Intelligence}, pages 1609--1617, 2024{\natexlab{a}}.

\bibitem[Dong et~al.(2024{\natexlab{b}})Dong, Zhang, and Zhang]{MAL24}
Yue-Jiang Dong, Fang-Lue Zhang, and Song-Hai Zhang.
\newblock Mal: Motion-aware loss with temporal and distillation hints for self-supervised depth estimation.
\newblock In \emph{2024 IEEE International Conference on Robotics and Automation (ICRA)}, pages 7318--7324, 2024{\natexlab{b}}.

\bibitem[Du et~al.(2020)Du, Turner, Dzitsiuk, Prasso, Duarte, Dourgarian, Afonso, Pascoal, Gladstone, Cruces, et~al.]{du2020depthlab}
Ruofei Du, Eric Turner, Maksym Dzitsiuk, Luca Prasso, Ivo Duarte, Jason Dourgarian, Joao Afonso, Jose Pascoal, Josh Gladstone, Nuno Cruces, et~al.
\newblock Depthlab: Real-time 3d interaction with depth maps for mobile augmented reality.
\newblock In \emph{Proceedings of the 33rd Annual ACM Symposium on User Interface Software and Technology}, pages 829--843, 2020.

\bibitem[Eigen et~al.(2014)Eigen, Puhrsch, and Fergus]{eigen2014depth}
David Eigen, Christian Puhrsch, and Rob Fergus.
\newblock Depth map prediction from a single image using a multi-scale deep network.
\newblock \emph{Advances in neural information processing systems}, 27, 2014.

\bibitem[El~Banani et~al.(2024)El~Banani, Raj, Maninis, Kar, Li, Rubinstein, Sun, Guibas, Johnson, and Jampani]{ElBanani2024CVPR}
Mohamed El~Banani, Amit Raj, Kevis-Kokitsi Maninis, Abhishek Kar, Yuanzhen Li, Michael Rubinstein, Deqing Sun, Leonidas Guibas, Justin Johnson, and Varun Jampani.
\newblock Probing the 3d awareness of visual foundation models.
\newblock In \emph{Proceedings of the IEEE/CVF Conference on Computer Vision and Pattern Recognition (CVPR)}, pages 21795--21806, 2024.

\bibitem[Fazli~Imam et~al.(2024)Fazli~Imam, Fedaku~Marew, Hassan, Fiaz, Fikri~Aji, and Cholakkal]{fazli2024clip}
Mohamed Fazli~Imam, Rufael Fedaku~Marew, Jameel Hassan, Mustansar Fiaz, Alham Fikri~Aji, and Hisham Cholakkal.
\newblock Clip meets dino for tuning zero-shot classifier using unlabeled image collections.
\newblock \emph{arXiv e-prints}, pages arXiv--2411, 2024.

\bibitem[Feng et~al.(2025)Feng, Zhang, Chen, Hu, Lu, and Ge]{dualpathdepth25}
Cheng Feng, Congxuan Zhang, Zhen Chen, Weiming Hu, Ke Lu, and Liyue Ge.
\newblock Self-supervised monocular depth estimation with dual-path encoders and offset field interpolation.
\newblock \emph{IEEE Transactions on Image Processing}, 34:\penalty0 939--954, 2025.

\bibitem[Feng et~al.(2022)Feng, Yang, Jing, Wang, Tian, and Li]{dyamicdepth22}
Ziyue Feng, Liang Yang, Longlong Jing, Haiyan Wang, YingLi Tian, and Bing Li.
\newblock Disentangling object motion and occlusion for unsupervised multi-frame monocular depth.
\newblock In \emph{Computer Vision -- ECCV 2022}, pages 228--244, Cham, 2022. Springer Nature Switzerland.

\bibitem[Fu et~al.(2018)Fu, Gong, Wang, Batmanghelich, and Tao]{fu2018deep}
Huan Fu, Mingming Gong, Chaohui Wang, Kayhan Batmanghelich, and Dacheng Tao.
\newblock Deep ordinal regression network for monocular depth estimation.
\newblock In \emph{Proceedings of the IEEE conference on computer vision and pattern recognition}, pages 2002--2011, 2018.

\bibitem[Geiger et~al.(2013)Geiger, Lenz, Stiller, and Urtasun]{KITTI}
Andreas Geiger, Philip Lenz, Christoph Stiller, and Raquel Urtasun.
\newblock Vision meets robotics: The kitti dataset.
\newblock \emph{The International Journal of Robotics Research}, 32\penalty0 (11):\penalty0 1231--1237, 2013.

\bibitem[Godard et~al.(2017)Godard, Mac~Aodha, and Brostow]{godard2017unsupervised}
Cl{\'e}ment Godard, Oisin Mac~Aodha, and Gabriel~J Brostow.
\newblock Unsupervised monocular depth estimation with left-right consistency.
\newblock In \emph{Proceedings of the IEEE conference on computer vision and pattern recognition}, pages 270--279, 2017.

\bibitem[Godard et~al.(2019)Godard, Mac~Aodha, Firman, and Brostow]{monodepthv2}
Cl{\'e}ment Godard, Oisin Mac~Aodha, Michael Firman, and Gabriel~J Brostow.
\newblock Digging into self-supervised monocular depth estimation.
\newblock In \emph{Proceedings of the IEEE/CVF international conference on computer vision}, pages 3828--3838, 2019.

\bibitem[Gonzalez~Bello et~al.(2024)Gonzalez~Bello, Moon, and Kim]{Quat24}
Juan~Luis Gonzalez~Bello, Jaeho Moon, and Munchurl Kim.
\newblock Self-supervised monocular depth estimation with positional shift depth variance and adaptive disparity quantization.
\newblock \emph{IEEE Transactions on Image Processing}, 33:\penalty0 2074--2089, 2024.

\bibitem[GonzalezBello and Kim(2020)]{gonzalezbello2020forget}
Juan~Luis GonzalezBello and Munchurl Kim.
\newblock Forget about the lidar: Self-supervised depth estimators with med probability volumes.
\newblock \emph{Advances in Neural Information Processing Systems}, 33:\penalty0 12626--12637, 2020.

\bibitem[Gu et~al.(2021)Gu, Lin, Kuo, and Cui]{gu2021open}
Xiuye Gu, Tsung-Yi Lin, Weicheng Kuo, and Yin Cui.
\newblock Open-vocabulary detection via vision and language knowledge distillation.
\newblock \emph{arXiv preprint arXiv:2104.13921}, 2021.

\bibitem[Guizilini et~al.(2020{\natexlab{a}})Guizilini, Ambrus, Pillai, Raventos, and Gaidon]{packnet}
Vitor Guizilini, Rares Ambrus, Sudeep Pillai, Allan Raventos, and Adrien Gaidon.
\newblock 3d packing for self-supervised monocular depth estimation.
\newblock In \emph{CVPR}, 2020{\natexlab{a}}.

\bibitem[Guizilini et~al.(2020{\natexlab{b}})Guizilini, Hou, Li, Ambrus, and Gaidon]{packnet-semguided}
Vitor Guizilini, Rui Hou, Jie Li, Rares Ambrus, and Adrien Gaidon.
\newblock Semantically-guided representation learning for self-supervised monocular depth.
\newblock In \emph{ICLR}, 2020{\natexlab{b}}.

\bibitem[Guizilini et~al.(2022)Guizilini, Ambruș, Chen, Zakharov, and Gaidon]{Depthformer22}
Vitor Guizilini, Rareș Ambruș, Dian Chen, Sergey Zakharov, and Adrien Gaidon.
\newblock Multi-frame self-supervised depth with transformers.
\newblock In \emph{Proceedings of the IEEE/CVF Conference on Computer Vision and Pattern Recognition (CVPR)}, pages 160--170, 2022.

\bibitem[Guizilini et~al.(2024)Guizilini, Tokmakov, Dave, and Ambrus]{guizilini2024grin}
Vitor Guizilini, Pavel Tokmakov, Achal Dave, and Rares Ambrus.
\newblock Grin: Zero-shot metric depth with pixel-level diffusion.
\newblock \emph{arXiv preprint arXiv:2409.09896}, 2024.

\bibitem[Han and Shen(2025)]{han2025high}
Wencheng Han and Jianbing Shen.
\newblock High-precision self-supervised monocular depth estimation with rich-resource prior.
\newblock In \emph{European Conference on Computer Vision}, pages 146--162. Springer, 2025.

\bibitem[Han et~al.(2022)Han, Yin, Jin, Dai, and Shen]{brnet22}
Wencheng Han, Junbo Yin, Xiaogang Jin, Xiangdong Dai, and Jianbing Shen.
\newblock Brnet: Exploring comprehensive features for monocular depth estimation.
\newblock In \emph{European Conference on Computer Vision}, pages 586--602. Springer, 2022.

\bibitem[He et~al.(2020)He, Fan, Wu, Xie, and Girshick]{moco}
Kaiming He, Haoqi Fan, Yuxin Wu, Saining Xie, and Ross Girshick.
\newblock Momentum contrast for unsupervised visual representation learning.
\newblock In \emph{Proceedings of the IEEE/CVF conference on computer vision and pattern recognition}, pages 9729--9738, 2020.

\bibitem[He et~al.(2022)He, Chen, Xie, Li, Doll{\'a}r, and Girshick]{mae}
Kaiming He, Xinlei Chen, Saining Xie, Yanghao Li, Piotr Doll{\'a}r, and Ross Girshick.
\newblock Masked autoencoders are scalable vision learners.
\newblock In \emph{Proceedings of the IEEE/CVF Conference on Computer Vision and Pattern Recognition}, pages 16000--16009, 2022.

\bibitem[Hoiem et~al.(2007)Hoiem, Efros, and Hebert]{hoiem2007recovering}
Derek Hoiem, Alexei~A Efros, and Martial Hebert.
\newblock Recovering surface layout from an image.
\newblock \emph{IJCV}, 2007.

\bibitem[Hu et~al.(2024{\natexlab{a}})Hu, Yin, Zhang, Cai, Long, Chen, Wang, Yu, Shen, and Shen]{metric3d24}
Mu Hu, Wei Yin, Chi Zhang, Zhipeng Cai, Xiaoxiao Long, Hao Chen, Kaixuan Wang, Gang Yu, Chunhua Shen, and Shaojie Shen.
\newblock Metric3d v2: A versatile monocular geometric foundation model for zero-shot metric depth and surface normal estimation.
\newblock \emph{arXiv preprint arXiv:2404.15506}, 2024{\natexlab{a}}.

\bibitem[Hu et~al.(2024{\natexlab{b}})Hu, Zhang, Zhang, Hai, Yu, and He]{hu2024learning}
Xueting Hu, Ce Zhang, Yi Zhang, Bowen Hai, Ke Yu, and Zhihai He.
\newblock Learning to adapt clip for few-shot monocular depth estimation.
\newblock In \emph{Proceedings of the IEEE/CVF Winter Conference on Applications of Computer Vision}, pages 5594--5603, 2024{\natexlab{b}}.

\bibitem[Huang et~al.(2021)Huang, Huang, Zhu, Yun, and Du]{huang2021bevdet}
Junjie Huang, Guan Huang, Zheng Zhu, Ye Yun, and Dalong Du.
\newblock Bevdet: High-performance multi-camera 3d object detection in bird-eye-view.
\newblock \emph{arXiv preprint arXiv:2112.11790}, 2021.

\bibitem[Ji et~al.(2021)Ji, Li, Bhanu, and Xu]{ji2021monoindoor}
Pan Ji, Runze Li, Bir Bhanu, and Yi Xu.
\newblock Monoindoor: Towards good practice of self-supervised monocular depth estimation for indoor environments.
\newblock In \emph{Proceedings of the IEEE/CVF International Conference on Computer Vision}, pages 12787--12796, 2021.

\bibitem[Jia et~al.(2021)Jia, Yang, Xia, Chen, Parekh, Pham, Le, Sung, Li, and Duerig]{align}
Chao Jia, Yinfei Yang, Ye Xia, Yi-Ting Chen, Zarana Parekh, Hieu Pham, Quoc Le, Yun-Hsuan Sung, Zhen Li, and Tom Duerig.
\newblock Scaling up visual and vision-language representation learning with noisy text supervision.
\newblock In \emph{International Conference on Machine Learning}, pages 4904--4916. PMLR, 2021.

\bibitem[Jia et~al.(2022)Jia, Tang, Chen, Cardie, Belongie, Hariharan, and Lim]{vpt}
Menglin Jia, Luming Tang, Bor-Chun Chen, Claire Cardie, Serge Belongie, Bharath Hariharan, and Ser-Nam Lim.
\newblock Visual prompt tuning.
\newblock In \emph{Computer Vision--ECCV 2022: 17th European Conference, Tel Aviv, Israel, October 23--27, 2022, Proceedings, Part XXXIII}, pages 709--727. Springer, 2022.

\bibitem[Jiang et~al.(2023)Jiang, Liu, Liu, Zhao, Zhang, Gao, Zhang, Li, and Xiong]{cliptodino23}
Dongsheng Jiang, Yuchen Liu, Songlin Liu, Jin'e Zhao, Hao Zhang, Zhen Gao, Xiaopeng Zhang, Jin Li, and Hongkai Xiong.
\newblock From clip to dino: Visual encoders shout in multi-modal large language models.
\newblock \emph{arXiv preprint arXiv:2310.08825}, 2023.

\bibitem[Jiao et~al.(2018)Jiao, Cao, Song, and Lau]{jiao2018look}
Jianbo Jiao, Ying Cao, Yibing Song, and Rynson Lau.
\newblock Look deeper into depth: Monocular depth estimation with semantic booster and attention-driven loss.
\newblock In \emph{Proceedings of the European conference on computer vision (ECCV)}, pages 53--69, 2018.

\bibitem[Johnston and Carneiro(2020)]{Johnston2020CVPR}
Adrian Johnston and Gustavo Carneiro.
\newblock Self-supervised monocular trained depth estimation using self-attention and discrete disparity volume.
\newblock In \emph{Proceedings of the IEEE/CVF Conference on Computer Vision and Pattern Recognition (CVPR)}, 2020.

\bibitem[Ke et~al.(2024)Ke, Obukhov, Huang, Metzger, Daudt, and Schindler]{Ke24}
Bingxin Ke, Anton Obukhov, Shengyu Huang, Nando Metzger, Rodrigo~Caye Daudt, and Konrad Schindler.
\newblock Repurposing diffusion-based image generators for monocular depth estimation.
\newblock In \emph{Proceedings of the IEEE/CVF Conference on Computer Vision and Pattern Recognition (CVPR)}, pages 9492--9502, 2024.

\bibitem[Khan et~al.(2021)Khan, Kim, and Tompkin]{khan2021differentiable}
Numair Khan, Min~H Kim, and James Tompkin.
\newblock Differentiable diffusion for dense depth estimation from multi-view images.
\newblock In \emph{Proceedings of the IEEE/CVF Conference on Computer Vision and Pattern Recognition}, pages 8912--8921, 2021.

\bibitem[Khattak et~al.(2023)Khattak, Rasheed, Maaz, Khan, and Khan]{maple}
Muhammad~Uzair Khattak, Hanoona Rasheed, Muhammad Maaz, Salman Khan, and Fahad~Shahbaz Khan.
\newblock Maple: Multi-modal prompt learning.
\newblock In \emph{Proceedings of the IEEE/CVF Conference on Computer Vision and Pattern Recognition}, pages 19113--19122, 2023.

\bibitem[Kim and Lee(2024)]{kim2024clip}
Dunam Kim and Seokju Lee.
\newblock Clip can understand depth.
\newblock \emph{arXiv preprint arXiv:2402.03251}, 2024.

\bibitem[Kim et~al.(2024)Kim, Pertsch, Karamcheti, Xiao, Balakrishna, Nair, Rafailov, Foster, Lam, Sanketi, Vuong, Kollar, Burchfiel, Tedrake, Sadigh, Levine, Liang, and Finn]{openvla}
{Moo Jin} Kim, Karl Pertsch, Siddharth Karamcheti, Ted Xiao, Ashwin Balakrishna, Suraj Nair, Rafael Rafailov, Ethan Foster, Grace Lam, Pannag Sanketi, Quan Vuong, Thomas Kollar, Benjamin Burchfiel, Russ Tedrake, Dorsa Sadigh, Sergey Levine, Percy Liang, and Chelsea Finn.
\newblock Openvla: An open-source vision-language-action model.
\newblock \emph{arXiv preprint arXiv:2406.09246}, 2024.

\bibitem[Klingner et~al.(2020)Klingner, Term\"{o}hlen, Mikolajczyk, and Fingscheidt]{klingner2020selfsupervised}
Marvin Klingner, Jan-Aike Term\"{o}hlen, Jonas Mikolajczyk, and Tim Fingscheidt.
\newblock {Self-Supervised Monocular Depth Estimation: Solving the Dynamic Object Problem by Semantic Guidance}.
\newblock In \emph{{European Conference on Computer Vision ({ECCV})}}, 2020.

\bibitem[Leduc et~al.(2024)Leduc, Cioppa, Giancola, Ghanem, and Van~Droogenbroeck]{soccernet24}
Arnaud Leduc, Anthony Cioppa, Silvio Giancola, Bernard Ghanem, and Marc Van~Droogenbroeck.
\newblock Soccernet-depth: a scalable dataset for monocular depth estimation in sports videos.
\newblock In \emph{Proceedings of the IEEE/CVF Conference on Computer Vision and Pattern Recognition}, pages 3280--3292, 2024.

\bibitem[Lester et~al.(2021)Lester, Al-Rfou, and Constant]{prompttuning}
Brian Lester, Rami Al-Rfou, and Noah Constant.
\newblock The power of scale for parameter-efficient prompt tuning.
\newblock \emph{arXiv preprint arXiv:2104.08691}, 2021.

\bibitem[Li et~al.(2023{\natexlab{a}})Li, Sun, Liang, Du, Zhang, Wang, Wang, Jin, and Zeng]{li2023bridging}
Bohan Li, Yasheng Sun, Zhujin Liang, Dalong Du, Zhuanghui Zhang, Xiaofeng Wang, Yunnan Wang, Xin Jin, and Wenjun Zeng.
\newblock Bridging stereo geometry and bev representation with reliable mutual interaction for semantic scene completion.
\newblock \emph{arXiv preprint arXiv:2303.13959}, 2023{\natexlab{a}}.

\bibitem[Li et~al.(2024{\natexlab{a}})Li, Deng, Zhang, Liang, Du, Jin, and Zeng]{li2024hierarchical}
Bohan Li, Jiajun Deng, Wenyao Zhang, Zhujin Liang, Dalong Du, Xin Jin, and Wenjun Zeng.
\newblock Hierarchical temporal context learning for camera-based semantic scene completion.
\newblock In \emph{European Conference on Computer Vision}, pages 131--148. Springer, 2024{\natexlab{a}}.

\bibitem[Li et~al.(2024{\natexlab{b}})Li, Guo, Liu, Zou, Ding, Chen, Zhu, Tan, Zhang, Wang, et~al.]{li2024uniscene}
Bohan Li, Jiazhe Guo, Hongsi Liu, Yingshuang Zou, Yikang Ding, Xiwu Chen, Hu Zhu, Feiyang Tan, Chi Zhang, Tiancai Wang, et~al.
\newblock Uniscene: Unified occupancy-centric driving scene generation.
\newblock \emph{arXiv preprint arXiv:2412.05435}, 2024{\natexlab{b}}.

\bibitem[Li et~al.(2024{\natexlab{c}})Li, Sun, Dong, Zhu, Liu, Jin, and Zeng]{li2024one}
Bohan Li, Yasheng Sun, Jingxin Dong, Zheng Zhu, Jinming Liu, Xin Jin, and Wenjun Zeng.
\newblock One at a time: Progressive multi-step volumetric probability learning for reliable 3d scene perception.
\newblock In \emph{Proceedings of the AAAI Conference on Artificial Intelligence}, pages 3028--3036, 2024{\natexlab{c}}.

\bibitem[Li et~al.(2022{\natexlab{a}})Li, Zhang, Zhang, Yang, Li, Zhong, Wang, Yuan, Zhang, Hwang, et~al.]{GLIP}
Liunian~Harold Li, Pengchuan Zhang, Haotian Zhang, Jianwei Yang, Chunyuan Li, Yiwu Zhong, Lijuan Wang, Lu Yuan, Lei Zhang, Jenq-Neng Hwang, et~al.
\newblock Grounded language-image pre-training.
\newblock In \emph{Proceedings of the IEEE/CVF Conference on Computer Vision and Pattern Recognition}, pages 10965--10975, 2022{\natexlab{a}}.

\bibitem[Li et~al.(2023{\natexlab{b}})Li, Sun, and Li]{li2023clip}
Siyuan Li, Li Sun, and Qingli Li.
\newblock Clip-reid: exploiting vision-language model for image re-identification without concrete text labels.
\newblock In \emph{Proceedings of the AAAI Conference on Artificial Intelligence}, pages 1405--1413, 2023{\natexlab{b}}.

\bibitem[Li and Snavely(2018)]{li2018megadepth}
Zhengqi Li and Noah Snavely.
\newblock Megadepth: Learning single-view depth prediction from internet photos.
\newblock In \emph{Proceedings of the IEEE conference on computer vision and pattern recognition}, pages 2041--2050, 2018.

\bibitem[Li et~al.(2022{\natexlab{b}})Li, Wang, Liu, and Jiang]{binsformer}
Zhenyu Li, Xuyang Wang, Xianming Liu, and Junjun Jiang.
\newblock Binsformer: Revisiting adaptive bins for monocular depth estimation.
\newblock \emph{arXiv:2204.00987}, 2022{\natexlab{b}}.

\bibitem[Li et~al.(2023{\natexlab{c}})Li, Yu, Wang, Anandkumar, Lu, and Alvarez]{li2023fbbev}
Zhiqi Li, Zhiding Yu, Wenhai Wang, Anima Anandkumar, Tong Lu, and Jose~M Alvarez.
\newblock {FB-BEV}: {BEV} representation from forward-backward view transformations.
\newblock In \emph{IEEE/CVF International Conference on Computer Vision (ICCV)}, 2023{\natexlab{c}}.

\bibitem[Liang et~al.(2023{\natexlab{a}})Liang, Xie, Zou, Ye, Xu, and Bai]{liang2023crowdclip}
Dingkang Liang, Jiahao Xie, Zhikang Zou, Xiaoqing Ye, Wei Xu, and Xiang Bai.
\newblock Crowdclip: Unsupervised crowd counting via vision-language model.
\newblock In \emph{Proceedings of the IEEE/CVF Conference on Computer Vision and Pattern Recognition}, pages 2893--2903, 2023{\natexlab{a}}.

\bibitem[Liang et~al.(2023{\natexlab{b}})Liang, Wu, Dai, Li, Zhao, Zhang, Zhang, Vajda, and Marculescu]{liang2023open}
Feng Liang, Bichen Wu, Xiaoliang Dai, Kunpeng Li, Yinan Zhao, Hang Zhang, Peizhao Zhang, Peter Vajda, and Diana Marculescu.
\newblock Open-vocabulary semantic segmentation with mask-adapted clip.
\newblock In \emph{Proceedings of the IEEE/CVF Conference on Computer Vision and Pattern Recognition}, pages 7061--7070, 2023{\natexlab{b}}.

\bibitem[Lin et~al.(2014)Lin, Maire, Belongie, Hays, Perona, Ramanan, Doll{\'a}r, and Zitnick]{coco}
Tsung-Yi Lin, Michael Maire, Serge Belongie, James Hays, Pietro Perona, Deva Ramanan, Piotr Doll{\'a}r, and C~Lawrence Zitnick.
\newblock Microsoft coco: Common objects in context.
\newblock In \emph{Computer Vision--ECCV 2014: 13th European Conference, Zurich, Switzerland, September 6-12, 2014, Proceedings, Part V 13}, pages 740--755. Springer, 2014.

\bibitem[Liu et~al.(2008)Liu, Yuen, Torralba, Sivic, and Freeman]{liu2008sift}
Ce Liu, Jenny Yuen, Antonio Torralba, Josef Sivic, and William~T Freeman.
\newblock Sift flow: Dense correspondence across different scenes.
\newblock In \emph{ECCV}, 2008.

\bibitem[Liu et~al.(2024)Liu, Kong, Li, Wang, Gu, and Chen]{liu2024mono}
Jinfeng Liu, Lingtong Kong, Bo Li, Zerong Wang, Hong Gu, and Jinwei Chen.
\newblock Mono-vifi: A unified learning framework for self-supervised single and multi-frame monocular depth estimation.
\newblock In \emph{European Conference on Computer Vision}, pages 90--107. Springer, 2024.

\bibitem[Lu et~al.(2022)Lu, Liu, Zhang, Liu, and Tian]{proda}
Yuning Lu, Jianzhuang Liu, Yonggang Zhang, Yajing Liu, and Xinmei Tian.
\newblock Prompt distribution learning.
\newblock In \emph{Proceedings of the IEEE/CVF Conference on Computer Vision and Pattern Recognition}, pages 5206--5215, 2022.

\bibitem[L\"uddecke and Ecker(2022)]{lueddecke22cvpr}
Timo L\"uddecke and Alexander Ecker.
\newblock Image segmentation using text and image prompts.
\newblock In \emph{Proceedings of the IEEE/CVF Conference on Computer Vision and Pattern Recognition (CVPR)}, pages 7086--7096, 2022.

\bibitem[Luo et~al.(2023)Luo, Bao, Wu, He, and Li]{Luo2023SegCLIP}
Huaishao Luo, Junwei Bao, Youzheng Wu, Xiaodong He, and Tianrui Li.
\newblock {SegCLIP}: Patch aggregation with learnable centers for open-vocabulary semantic segmentation.
\newblock \emph{ICML}, 2023.

\bibitem[Lyu et~al.(2021)Lyu, Liu, Wang, Kong, Liu, Liu, Chen, and Yuan]{lyu2020hr}
Xiaoyang Lyu, Liang Liu, Mengmeng Wang, Xin Kong, Lina Liu, Yong Liu, Xinxin Chen, and Yi Yuan.
\newblock Hr-depth: High resolution self-supervised monocular depth estimation.
\newblock In \emph{Proceedings of the AAAI Conference on Artificial Intelligence}, pages 2294--2301, 2021.

\bibitem[Nguyen et~al.(2024)Nguyen, Wang, Alvarez, and Liu]{Nguyen2024CVPR}
Hoang~Chuong Nguyen, Tianyu Wang, Jose~M. Alvarez, and Miaomiao Liu.
\newblock Mining supervision for dynamic regions in self-supervised monocular depth estimation.
\newblock In \emph{Proceedings of the IEEE/CVF Conference on Computer Vision and Pattern Recognition (CVPR)}, pages 10446--10455, 2024.

\bibitem[Okae et~al.(2021)Okae, Li, Du, and Hu]{okae2021robust}
James Okae, Bohan Li, Juan Du, and Yueming Hu.
\newblock Robust scale-aware stereo matching network.
\newblock \emph{IEEE Transactions on Artificial Intelligence}, 3\penalty0 (2):\penalty0 244--253, 2021.

\bibitem[Oquab et~al.(2023)Oquab, Darcet, Moutakanni, Vo, Szafraniec, Khalidov, Fernandez, Haziza, Massa, El-Nouby, et~al.]{dinov223}
Maxime Oquab, Timoth{\'e}e Darcet, Th{\'e}o Moutakanni, Huy Vo, Marc Szafraniec, Vasil Khalidov, Pierre Fernandez, Daniel Haziza, Francisco Massa, Alaaeldin El-Nouby, et~al.
\newblock Dinov2: Learning robust visual features without supervision.
\newblock \emph{arXiv preprint arXiv:2304.07193}, 2023.

\bibitem[Park et~al.(2024)Park, Jeong, Lee, and Jeon]{Park24}
Jin-Hwi Park, Chanhwi Jeong, Junoh Lee, and Hae-Gon Jeon.
\newblock Depth prompting for sensor-agnostic depth estimation.
\newblock In \emph{Proceedings of the IEEE/CVF Conference on Computer Vision and Pattern Recognition (CVPR)}, pages 9859--9869, 2024.

\bibitem[Paszke et~al.(2017)Paszke, Gross, Chintala, Chanan, Yang, DeVito, Lin, Desmaison, Antiga, and Lerer]{pytorch}
Adam Paszke, Sam Gross, Soumith Chintala, Gregory Chanan, Edward Yang, Zachary DeVito, Zeming Lin, Alban Desmaison, Luca Antiga, and Adam Lerer.
\newblock Automatic differentiation in pytorch.
\newblock 2017.

\bibitem[Patni et~al.(2024)Patni, Agarwal, and Arora]{ecodepth24}
Suraj Patni, Aradhye Agarwal, and Chetan Arora.
\newblock Ecodepth: Effective conditioning of diffusion models for monocular depth estimation.
\newblock In \emph{Proceedings of the IEEE/CVF Conference on Computer Vision and Pattern Recognition}, pages 28285--28295, 2024.

\bibitem[Petroni et~al.(2019)Petroni, Rockt{\"a}schel, Lewis, Bakhtin, Wu, Miller, and Riedel]{petroni2019language}
Fabio Petroni, Tim Rockt{\"a}schel, Patrick Lewis, Anton Bakhtin, Yuxiang Wu, Alexander~H Miller, and Sebastian Riedel.
\newblock Language models as knowledge bases?
\newblock \emph{arXiv preprint arXiv:1909.01066}, 2019.

\bibitem[Pham et~al.(2024)Pham, Do, Nguyen, Hua, Nguyen, and Nguyen]{sharpdepth25}
Duc-Hai Pham, Tung Do, Phong Nguyen, Binh-Son Hua, Khoi Nguyen, and Rang Nguyen.
\newblock Sharpdepth: Sharpening metric depth predictions using diffusion distillation.
\newblock \emph{arXiv preprint arXiv:2411.18229}, 2024.

\bibitem[Piccinelli et~al.(2024)Piccinelli, Yang, Sakaridis, Segu, Li, Van~Gool, and Yu]{Piccinelli2024CVPR}
Luigi Piccinelli, Yung-Hsu Yang, Christos Sakaridis, Mattia Segu, Siyuan Li, Luc Van~Gool, and Fisher Yu.
\newblock Unidepth: Universal monocular metric depth estimation.
\newblock In \emph{Proceedings of the IEEE/CVF Conference on Computer Vision and Pattern Recognition (CVPR)}, pages 10106--10116, 2024.

\bibitem[Poggi et~al.(2020)Poggi, Aleotti, Tosi, and Mattoccia]{poggi2020uncertainty}
Matteo Poggi, Filippo Aleotti, Fabio Tosi, and Stefano Mattoccia.
\newblock On the uncertainty of self-supervised monocular depth estimation.
\newblock In \emph{Proceedings of the IEEE/CVF Conference on Computer Vision and Pattern Recognition}, pages 3227--3237, 2020.

\bibitem[Qi et~al.(2023)Qi, Dong, Fan, Ge, Zhang, Ma, and Yi]{recon23}
Zekun Qi, Runpei Dong, Guofan Fan, Zheng Ge, Xiangyu Zhang, Kaisheng Ma, and Li Yi.
\newblock Contrast with reconstruct: Contrastive 3d representation learning guided by generative pretraining.
\newblock In \emph{International Conference on Machine Learning}, pages 28223--28243. PMLR, 2023.

\bibitem[Qi et~al.(2024)Qi, Dong, Zhang, Geng, Han, Ge, Yi, and Ma]{shapellm24}
Zekun Qi, Runpei Dong, Shaochen Zhang, Haoran Geng, Chunrui Han, Zheng Ge, Li Yi, and Kaisheng Ma.
\newblock Shapellm: Universal 3d object understanding for embodied interaction.
\newblock In \emph{European Conference on Computer Vision}, pages 214--238. Springer, 2024.

\bibitem[Qi et~al.(2025)Qi, Zhang, Ding, Dong, Yu, Li, Xu, Li, He, Fan, et~al.]{sofar25}
Zekun Qi, Wenyao Zhang, Yufei Ding, Runpei Dong, Xinqiang Yu, Jingwen Li, Lingyun Xu, Baoyu Li, Xialin He, Guofan Fan, et~al.
\newblock Sofar: Language-grounded orientation bridges spatial reasoning and object manipulation.
\newblock \emph{arXiv preprint arXiv:2502.13143}, 2025.

\bibitem[Radford et~al.(2021)Radford, Kim, Hallacy, Ramesh, Goh, Agarwal, Sastry, Askell, Mishkin, Clark, et~al.]{clip}
Alec Radford, Jong~Wook Kim, Chris Hallacy, Aditya Ramesh, Gabriel Goh, Sandhini Agarwal, Girish Sastry, Amanda Askell, Pamela Mishkin, Jack Clark, et~al.
\newblock Learning transferable visual models from natural language supervision.
\newblock In \emph{International conference on machine learning}, pages 8748--8763. PMLR, 2021.

\bibitem[Ranftl et~al.(2021)Ranftl, Bochkovskiy, and Koltun]{dpt21}
Ren{\'e} Ranftl, Alexey Bochkovskiy, and Vladlen Koltun.
\newblock Vision transformers for dense prediction.
\newblock In \emph{Proceedings of the IEEE/CVF international conference on computer vision}, pages 12179--12188, 2021.

\bibitem[Ranftl et~al.(2022)Ranftl, Lasinger, Hafner, Schindler, and Koltun]{midas22}
Ren\'{e} Ranftl, Katrin Lasinger, David Hafner, Konrad Schindler, and Vladlen Koltun.
\newblock Towards robust monocular depth estimation: Mixing datasets for zero-shot cross-dataset transfer.
\newblock \emph{IEEE Transactions on Pattern Analysis and Machine Intelligence}, 44\penalty0 (3), 2022.

\bibitem[Rao et~al.(2022)Rao, Zhao, Chen, Tang, Zhu, Huang, Zhou, and Lu]{denseclip}
Yongming Rao, Wenliang Zhao, Guangyi Chen, Yansong Tang, Zheng Zhu, Guan Huang, Jie Zhou, and Jiwen Lu.
\newblock Denseclip: Language-guided dense prediction with context-aware prompting.
\newblock In \emph{Proceedings of the IEEE/CVF Conference on Computer Vision and Pattern Recognition}, pages 18082--18091, 2022.

\bibitem[Saunders et~al.(2023)Saunders, Vogiatzis, and Manso]{saunders2023self}
Kieran Saunders, George Vogiatzis, and Luis~J Manso.
\newblock Self-supervised monocular depth estimation: Let's talk about the weather.
\newblock In \emph{Proceedings of the IEEE/CVF International Conference on Computer Vision}, pages 8907--8917, 2023.

\bibitem[Saxena et~al.(2008)Saxena, Sun, and Ng]{make3d}
Ashutosh Saxena, Min Sun, and Andrew~Y Ng.
\newblock Make3d: Learning 3d scene structure from a single still image.
\newblock \emph{TPAMI}, 2008.

\bibitem[Saxena et~al.(2024)Saxena, Herrmann, Hur, Kar, Norouzi, Sun, and Fleet]{surprising24}
Saurabh Saxena, Charles Herrmann, Junhwa Hur, Abhishek Kar, Mohammad Norouzi, Deqing Sun, and David~J Fleet.
\newblock The surprising effectiveness of diffusion models for optical flow and monocular depth estimation.
\newblock \emph{Advances in Neural Information Processing Systems}, 36, 2024.

\bibitem[Shao et~al.(2023)Shao, Pei, Chen, Wu, and Li]{shao2023nddepth}
Shuwei Shao, Zhongcai Pei, Weihai Chen, Xingming Wu, and Zhengguo Li.
\newblock Nddepth: Normal-distance assisted monocular depth estimation.
\newblock In \emph{Proceedings of the IEEE/CVF International Conference on Computer Vision}, pages 7931--7940, 2023.

\bibitem[Shao et~al.(2024)Shao, Pei, Chen, Sun, Chen, and Li]{monodiffusion24}
Shuwei Shao, Zhongcai Pei, Weihai Chen, Dingchi Sun, Peter~CY Chen, and Zhengguo Li.
\newblock Monodiffusion: self-supervised monocular depth estimation using diffusion model.
\newblock \emph{IEEE Transactions on Circuits and Systems for Video Technology}, 2024.

\bibitem[Shu et~al.(2020)Shu, Yu, Duan, and Yang]{shu2020feature}
Chang Shu, Kun Yu, Zhixiang Duan, and Kuiyuan Yang.
\newblock Feature-metric loss for self-supervised learning of depth and egomotion.
\newblock In \emph{European Conference on Computer Vision}, pages 572--588. Springer, 2020.

\bibitem[Si et~al.(2023)Si, Zhao, Wang, Gao, Chen, Wang, and Li]{Si2023CVPR}
Haozhe Si, Bin Zhao, Dong Wang, Yunpeng Gao, Mulin Chen, Zhigang Wang, and Xuelong Li.
\newblock Fully self-supervised depth estimation from defocus clue.
\newblock In \emph{Proceedings of the IEEE/CVF Conference on Computer Vision and Pattern Recognition (CVPR)}, pages 9140--9149, 2023.

\bibitem[Son and Lee(2024)]{clipcabins24}
Eunjin Son and Sang~Jun Lee.
\newblock Cabins: Clip-based adaptive bins for monocular depth estimation.
\newblock In \emph{Proceedings of the IEEE/CVF Conference on Computer Vision and Pattern Recognition}, pages 4557--4567, 2024.

\bibitem[Stevens et~al.(2025)Stevens, Chao, Berger-Wolf, and Su]{sparse25}
Samuel Stevens, Wei-Lun Chao, Tanya Berger-Wolf, and Yu Su.
\newblock Sparse autoencoders for scientifically rigorous interpretation of vision models.
\newblock \emph{arXiv preprint arXiv:2502.06755}, 2025.

\bibitem[Tong et~al.(2024)Tong, Liu, Zhai, Ma, LeCun, and Xie]{tong2024eyes}
Shengbang Tong, Zhuang Liu, Yuexiang Zhai, Yi Ma, Yann LeCun, and Saining Xie.
\newblock Eyes wide shut? exploring the visual shortcomings of multimodal llms.
\newblock In \emph{Proceedings of the IEEE/CVF Conference on Computer Vision and Pattern Recognition}, pages 9568--9578, 2024.

\bibitem[Vankadari et~al.(2020)Vankadari, Garg, Majumder, Kumar, and Behera]{vankadari2020unsupervised}
Madhu Vankadari, Sourav Garg, Anima Majumder, Swagat Kumar, and Ardhendu Behera.
\newblock Unsupervised monocular depth estimation for night-time images using adversarial domain feature adaptation.
\newblock In \emph{Computer Vision--ECCV 2020: 16th European Conference, Glasgow, UK, August 23--28, 2020, Proceedings, Part XXVIII 16}, pages 443--459. Springer, 2020.

\bibitem[Vankadari et~al.(2024)Vankadari, Hodgson, Shin, Zhou, Markham, and Trigoni]{10610318}
Madhu Vankadari, Samuel Hodgson, Sangyun Shin, Kaichen Zhou, Andrew Markham, and Niki Trigoni.
\newblock Dusk till dawn: Self-supervised nighttime stereo depth estimation using visual foundation models.
\newblock In \emph{2024 IEEE International Conference on Robotics and Automation (ICRA)}, pages 17976--17982, 2024.

\bibitem[Walia et~al.(2022)Walia, Walz, Bijelic, Mannan, Julca-Aguilar, Langer, Ritter, and Heide]{Walia2022CVPR}
Amanpreet Walia, Stefanie Walz, Mario Bijelic, Fahim Mannan, Frank Julca-Aguilar, Michael Langer, Werner Ritter, and Felix Heide.
\newblock Gated2gated: Self-supervised depth estimation from gated images.
\newblock In \emph{Proceedings of the IEEE/CVF Conference on Computer Vision and Pattern Recognition (CVPR)}, pages 2811--2821, 2022.

\bibitem[Wang and Liu(2024)]{wangdepth24}
Ning-Hsu Wang and Yu-Lun Liu.
\newblock Depth anywhere: Enhancing 360 monocular depth estimation via perspective distillation and unlabeled data augmentation.
\newblock \emph{arXiv preprint arXiv:2406.12849}, 2024.

\bibitem[Wang et~al.(2023)Wang, Yu, and Gao]{wang2023planedepth}
Ruoyu Wang, Zehao Yu, and Shenghua Gao.
\newblock Planedepth: Self-supervised depth estimation via orthogonal planes.
\newblock In \emph{Proceedings of the IEEE/CVF Conference on Computer Vision and Pattern Recognition}, pages 21425--21434, 2023.

\bibitem[Wang et~al.(2019)Wang, Chao, Garg, Hariharan, Campbell, and Weinberger]{wang2019pseudo}
Yan Wang, Wei-Lun Chao, Divyansh Garg, Bharath Hariharan, Mark Campbell, and Kilian~Q Weinberger.
\newblock Pseudo-lidar from visual depth estimation: Bridging the gap in 3d object detection for autonomous driving.
\newblock In \emph{Proceedings of the IEEE/CVF Conference on Computer Vision and Pattern Recognition}, pages 8445--8453, 2019.

\bibitem[Wang et~al.(2022)Wang, Wang, Liang, Yang, An, and Guo]{Wang2022CVPR}
Yingqian Wang, Longguang Wang, Zhengyu Liang, Jungang Yang, Wei An, and Yulan Guo.
\newblock Occlusion-aware cost constructor for light field depth estimation.
\newblock In \emph{Proceedings of the IEEE/CVF Conference on Computer Vision and Pattern Recognition (CVPR)}, pages 19809--19818, 2022.

\bibitem[Wang et~al.(2024)Wang, Liang, Xu, Jiao, and Yu]{WangLiangXuJiaoYu2024}
Youhong Wang, Yunji Liang, Hao Xu, Shaohui Jiao, and Hongkai Yu.
\newblock Sqldepth: Generalizable self-supervised fine-structured monocular depth estimation.
\newblock \emph{Proceedings of the AAAI Conference on Artificial Intelligence}, 38\penalty0 (6):\penalty0 5713--5721, 2024.

\bibitem[Wang et~al.(2004)Wang, Bovik, Sheikh, and Simoncelli]{wang2004image}
Zhou Wang, Alan~C Bovik, Hamid~R Sheikh, and Eero~P Simoncelli.
\newblock Image quality assessment: from error visibility to structural similarity.
\newblock \emph{IEEE transactions on image processing}, 13\penalty0 (4):\penalty0 600--612, 2004.

\bibitem[Watson et~al.(2019)Watson, Firman, Brostow, and Turmukhambetov]{watson2019self}
Jamie Watson, Michael Firman, Gabriel~J Brostow, and Daniyar Turmukhambetov.
\newblock Self-supervised monocular depth hints.
\newblock In \emph{Proceedings of the IEEE/CVF International Conference on Computer Vision}, pages 2162--2171, 2019.

\bibitem[Watson et~al.(2021)Watson, Aodha, Prisacariu, Brostow, and Firman]{manydepth21}
Jamie Watson, Oisin~Mac Aodha, Victor Prisacariu, Gabriel Brostow, and Michael Firman.
\newblock {The Temporal Opportunist: Self-Supervised Multi-Frame Monocular Depth}.
\newblock In \emph{Computer Vision and Pattern Recognition (CVPR)}, 2021.

\bibitem[Wei et~al.(2024)Wei, Geng, Chen, Deng, Wenbo, Zhao, Fang, Guibas, and Wang]{wei2024d}
Songlin Wei, Haoran Geng, Jiayi Chen, Congyue Deng, Cui Wenbo, Chengyang Zhao, Xiaomeng Fang, Leonidas Guibas, and He Wang.
\newblock D3roma: Disparity diffusion-based depth sensing for material-agnostic robotic manipulation.
\newblock In \emph{ECCV 2024 Workshop on Wild 3D: 3D Modeling, Reconstruction, and Generation in the Wild}, 2024.

\bibitem[Wofk et~al.(2019)Wofk, Ma, Yang, Karaman, and Sze]{wofk2019fastdepth}
Diana Wofk, Fangchang Ma, Tien-Ju Yang, Sertac Karaman, and Vivienne Sze.
\newblock Fastdepth: Fast monocular depth estimation on embedded systems.
\newblock In \emph{2019 International Conference on Robotics and Automation (ICRA)}, pages 6101--6108. IEEE, 2019.

\bibitem[Xian et~al.(2018)Xian, Shen, Cao, Lu, Xiao, Li, and Luo]{redweb}
Ke Xian, Chunhua Shen, Zhiguo Cao, Hao Lu, Yang Xiao, Ruibo Li, and Zhenbo Luo.
\newblock Monocular relative depth perception with web stereo data supervision.
\newblock In \emph{CVPR}, 2018.

\bibitem[Yang et~al.(2024{\natexlab{a}})Yang, Kang, Huang, Xu, Feng, and Zhao]{depthanything}
Lihe Yang, Bingyi Kang, Zilong Huang, Xiaogang Xu, Jiashi Feng, and Hengshuang Zhao.
\newblock Depth anything: Unleashing the power of large-scale unlabeled data.
\newblock In \emph{Proceedings of the IEEE/CVF Conference on Computer Vision and Pattern Recognition}, pages 10371--10381, 2024{\natexlab{a}}.

\bibitem[Yang et~al.(2024{\natexlab{b}})Yang, Kang, Huang, Zhao, Xu, Feng, and Zhao]{depthanythingv2}
Lihe Yang, Bingyi Kang, Zilong Huang, Zhen Zhao, Xiaogang Xu, Jiashi Feng, and Hengshuang Zhao.
\newblock Depth anything v2.
\newblock \emph{arXiv:2406.09414}, 2024{\natexlab{b}}.

\bibitem[Yang et~al.(2024{\natexlab{c}})Yang, Pan, Dai, Sun, and Xiao]{YANG2024106410}
Zhuoyue Yang, Junjun Pan, Ju Dai, Zhen Sun, and Yi Xiao.
\newblock Self-supervised endoscopy depth estimation framework with clip-guidance segmentation.
\newblock \emph{Biomedical Signal Processing and Control}, 95:\penalty0 106410, 2024{\natexlab{c}}.

\bibitem[Yao et~al.(2021)Yao, Zhang, Zhang, Liu, Chua, and Sun]{yao2021cpt}
Yuan Yao, Ao Zhang, Zhengyan Zhang, Zhiyuan Liu, Tat-Seng Chua, and Maosong Sun.
\newblock Cpt: Colorful prompt tuning for pre-trained vision-language models.
\newblock \emph{arXiv preprint arXiv:2109.11797}, 2021.

\bibitem[Yin et~al.(2019)Yin, Liu, Shen, and Yan]{vnl}
Wei Yin, Yifan Liu, Chunhua Shen, and Youliang Yan.
\newblock Enforcing geometric constraints of virtual normal for depth prediction.
\newblock In \emph{ICCV}, 2019.

\bibitem[Yin et~al.(2023)Yin, Zhang, Chen, Cai, Yu, Wang, Chen, and Shen]{metric3d23}
Wei Yin, Chi Zhang, Hao Chen, Zhipeng Cai, Gang Yu, Kaixuan Wang, Xiaozhi Chen, and Chunhua Shen.
\newblock Metric3d: Towards zero-shot metric 3d prediction from a single image.
\newblock In \emph{Proceedings of the IEEE/CVF International Conference on Computer Vision}, pages 9043--9053, 2023.

\bibitem[Yin and Shi(2018)]{yin2018geonet}
Zhichao Yin and Jianping Shi.
\newblock {GeoNet}: Unsupervised learning of dense depth, optical flow and camera pose.
\newblock In \emph{CVPR}, 2018.

\bibitem[You et~al.(2020)You, Wang, Chao, Garg, Pleiss, Hariharan, Campbell, and Weinberger]{you2020pseudo}
Yurong You, Yan Wang, Wei-Lun Chao, Divyansh Garg, Geoff Pleiss, Bharath Hariharan, Mark Campbell, and Kilian~Q Weinberger.
\newblock Pseudo-lidar++: Accurate depth for 3d object detection in autonomous driving.
\newblock In \emph{ICLR}, 2020.

\bibitem[Zeng et~al.(2024{\natexlab{a}})Zeng, Ni, Wang, Rim, Chung, Yang, Hong, and Wong]{zeng2024priordiffusion}
Ziyao Zeng, Jingcheng Ni, Daniel Wang, Patrick Rim, Younjoon Chung, Fengyu Yang, Byung-Woo Hong, and Alex Wong.
\newblock Priordiffusion: Leverage language prior in diffusion models for monocular depth estimation.
\newblock \emph{arXiv preprint arXiv:2411.16750}, 2024{\natexlab{a}}.

\bibitem[Zeng et~al.(2024{\natexlab{b}})Zeng, Wang, Yang, Park, Soatto, Lao, and Wong]{zeng2024wordepth}
Ziyao Zeng, Daniel Wang, Fengyu Yang, Hyoungseob Park, Stefano Soatto, Dong Lao, and Alex Wong.
\newblock Wordepth: Variational language prior for monocular depth estimation.
\newblock In \emph{Proceedings of the IEEE/CVF Conference on Computer Vision and Pattern Recognition}, pages 9708--9719, 2024{\natexlab{b}}.

\bibitem[Zeng et~al.(2024{\natexlab{c}})Zeng, Wu, Park, Wang, Yang, Soatto, Lao, Hong, and Wong]{rsa24}
Ziyao Zeng, Yangchao Wu, Hyoungseob Park, Daniel Wang, Fengyu Yang, Stefano Soatto, Dong Lao, Byung-Woo Hong, and Alex Wong.
\newblock Rsa: Resolving scale ambiguities in monocular depth estimators through language descriptions.
\newblock \emph{arXiv preprint arXiv:2410.02924}, 2024{\natexlab{c}}.

\bibitem[Zhang et~al.(2025{\natexlab{a}})Zhang, Yu, Xiao, Wei, and Zhao]{10891864}
Bingfeng Zhang, Siyue Yu, Jimin Xiao, Yunchao Wei, and Yao Zhao.
\newblock Frozen clip-dino: a strong backbone for weakly supervised semantic segmentation.
\newblock \emph{IEEE Transactions on Pattern Analysis and Machine Intelligence}, pages 1--17, 2025{\natexlab{a}}.

\bibitem[Zhang et~al.(2023)Zhang, Nex, Vosselman, and Kerle]{litemono23}
Ning Zhang, Francesco Nex, George Vosselman, and Norman Kerle.
\newblock Lite-mono: A lightweight cnn and transformer architecture for self-supervised monocular depth estimation.
\newblock In \emph{Proceedings of the IEEE/CVF Conference on Computer Vision and Pattern Recognition (CVPR)}, pages 18537--18546, 2023.

\bibitem[Zhang et~al.(2022{\natexlab{a}})Zhang, Guo, Zhang, Li, Miao, Cui, Qiao, Gao, and Li]{zhang2022pointclip}
Renrui Zhang, Ziyu Guo, Wei Zhang, Kunchang Li, Xupeng Miao, Bin Cui, Yu Qiao, Peng Gao, and Hongsheng Li.
\newblock Pointclip: Point cloud understanding by clip.
\newblock In \emph{Proceedings of the IEEE/CVF Conference on Computer Vision and Pattern Recognition}, pages 8552--8562, 2022{\natexlab{a}}.

\bibitem[Zhang et~al.(2022{\natexlab{b}})Zhang, Zeng, Guo, and Li]{depthclip}
Renrui Zhang, Ziyao Zeng, Ziyu Guo, and Yafeng Li.
\newblock Can language understand depth?
\newblock In \emph{Proceedings of the 30th ACM International Conference on Multimedia}, pages 6868--6874, 2022{\natexlab{b}}.

\bibitem[Zhang et~al.(2022{\natexlab{c}})Zhang, Lyu, Xue, Yao, Zhu, and Jia]{zhang2022predict}
Wenyao Zhang, Shipeng Lyu, Feng Xue, Chen Yao, Zheng Zhu, and Zhenzhong Jia.
\newblock Predict the rover mobility over soft terrain using articulated wheeled bevameter.
\newblock \emph{IEEE Robotics and Automation Letters}, 7\penalty0 (4):\penalty0 12062--12069, 2022{\natexlab{c}}.

\bibitem[Zhang et~al.(2022{\natexlab{d}})Zhang, Lyu, Yao, Xue, Zhu, and Jia]{zhang2022analysis}
Wenyao Zhang, Shipeng Lyu, Chen Yao, Feng Xue, Zheng Zhu, and Zhenzhong Jia.
\newblock Analysis of robot traversability over unstructured terrain using information fusion.
\newblock In \emph{2022 International Conference on Advanced Robotics and Mechatronics (ICARM)}, pages 413--418. IEEE, 2022{\natexlab{d}}.

\bibitem[Zhang et~al.(2024)Zhang, Wu, Zhang, Yu, Ma, Jin, Yang, and Zeng]{wenyao24}
Wenyao Zhang, Letian Wu, Zequn Zhang, Tao Yu, Chao Ma, Xin Jin, Xiaokang Yang, and Wenjun Zeng.
\newblock Unleash the power of vision-language models by visual attention prompt and multi-modal interaction.
\newblock \emph{IEEE Transactions on Multimedia}, pages 1--13, 2024.

\bibitem[Zhang et~al.(2025{\natexlab{b}})Zhang, Liu, Qi, Wang, Yu, Zhang, Dong, He, Wang, Zhang, et~al.]{zhang2025dreamvla}
Wenyao Zhang, Hongsi Liu, Zekun Qi, Yunnan Wang, XinQiang Yu, Jiazhao Zhang, Runpei Dong, Jiawei He, He Wang, Zhizheng Zhang, et~al.
\newblock Dreamvla: A vision-language-action model dreamed with comprehensive world knowledge.
\newblock \emph{arXiv preprint arXiv:2507.04447}, 2025{\natexlab{b}}.

\bibitem[Zhao et~al.(2022)Zhao, Zhang, Poggi, Tosi, Guo, Zhu, Huang, Tang, and Mattoccia]{zhao2022monovit}
Chaoqiang Zhao, Youmin Zhang, Matteo Poggi, Fabio Tosi, Xianda Guo, Zheng Zhu, Guan Huang, Yang Tang, and Stefano Mattoccia.
\newblock Monovit: Self-supervised monocular depth estimation with a vision transformer.
\newblock In \emph{2022 International Conference on 3D Vision (3DV)}, pages 668--678. IEEE, 2022.

\bibitem[Zhao et~al.(2023)Zhao, Poggi, Tosi, Zhou, Sun, Tang, and Mattoccia]{zhao2023gasmono}
Chaoqiang Zhao, Matteo Poggi, Fabio Tosi, Lei Zhou, Qiyu Sun, Yang Tang, and Stefano Mattoccia.
\newblock Gasmono: Geometry-aided self-supervised monocular depth estimation for indoor scenes.
\newblock In \emph{Proceedings of the IEEE/CVF International Conference on Computer Vision}, pages 16209--16220, 2023.

\bibitem[Zhao et~al.(2016)Zhao, Gallo, Frosio, and Kautz]{zhao2016loss}
Hang Zhao, Orazio Gallo, Iuri Frosio, and Jan Kautz.
\newblock Loss functions for image restoration with neural networks.
\newblock \emph{IEEE Transactions on computational imaging}, 3\penalty0 (1):\penalty0 47--57, 2016.

\bibitem[Zhong et~al.(2022)Zhong, Yang, Zhang, Li, Codella, Li, Zhou, Dai, Yuan, Li, et~al.]{zhong2022regionclip}
Yiwu Zhong, Jianwei Yang, Pengchuan Zhang, Chunyuan Li, Noel Codella, Liunian~Harold Li, Luowei Zhou, Xiyang Dai, Lu Yuan, Yin Li, et~al.
\newblock Regionclip: Region-based language-image pretraining.
\newblock In \emph{Proceedings of the IEEE/CVF Conference on Computer Vision and Pattern Recognition}, pages 16793--16803, 2022.

\bibitem[Zhou et~al.(2021)Zhou, Greenwood, and Taylor]{DIFF21}
Hang Zhou, David Greenwood, and Sarah Taylor.
\newblock Self-supervised monocular depth estimation with internal feature fusion.
\newblock In \emph{British Machine Vision Conference (BMVC)}, 2021.

\bibitem[Zhou et~al.(2022{\natexlab{a}})Zhou, Yang, Loy, and Liu]{COCOOP}
Kaiyang Zhou, Jingkang Yang, Chen~Change Loy, and Ziwei Liu.
\newblock Conditional prompt learning for vision-language models.
\newblock In \emph{Proceedings of the IEEE/CVF Conference on Computer Vision and Pattern Recognition}, pages 16816--16825, 2022{\natexlab{a}}.

\bibitem[Zhou et~al.(2022{\natexlab{b}})Zhou, Yang, Loy, and Liu]{COOP}
Kaiyang Zhou, Jingkang Yang, Chen~Change Loy, and Ziwei Liu.
\newblock Learning to prompt for vision-language models.
\newblock \emph{International Journal of Computer Vision}, 130\penalty0 (9):\penalty0 2337--2348, 2022{\natexlab{b}}.

\bibitem[Zhou et~al.(2017)Zhou, Brown, Snavely, and Lowe]{zhou2017}
Tinghui Zhou, Matthew Brown, Noah Snavely, and David~G Lowe.
\newblock Unsupervised learning of depth and ego-motion from video.
\newblock In \emph{Proceedings of the IEEE conference on computer vision and pattern recognition}, pages 1851--1858, 2017.

\bibitem[Zhu et~al.(2022)Zhu, Yao, Zhu, Liu, and Jia]{zhuhu}
Hu Zhu, Chen Yao, Zheng Zhu, Zhengtao Liu, and Zhenzhong Jia.
\newblock Fusing panoptic segmentation and geometry information for robust visual slam in dynamic environments.
\newblock In \emph{2022 IEEE 18th International Conference on Automation Science and Engineering (CASE)}, pages 1648--1653, 2022.

\bibitem[Zhu et~al.(2023{\natexlab{a}})Zhu, Li, Chen, Fan, Mao, Jing, Liu, and Shen]{zhu2023segprompt}
Muzhi Zhu, Hengtao Li, Hao Chen, Chengxiang Fan, Weian Mao, Chenchen Jing, Yifan Liu, and Chunhua Shen.
\newblock Segprompt: Boosting open-world segmentation via category-level prompt learning.
\newblock In \emph{Proceedings of the IEEE/CVF International Conference on Computer Vision}, pages 999--1008, 2023{\natexlab{a}}.

\bibitem[Zhu et~al.(2020)Zhu, Brazil, and Liu]{zhu2020edge}
Shengjie Zhu, Garrick Brazil, and Xiaoming Liu.
\newblock The edge of depth: Explicit constraints between segmentation and depth.
\newblock In \emph{Proceedings of the IEEE/CVF conference on computer vision and pattern recognition}, pages 13116--13125, 2020.

\bibitem[Zhu et~al.(2023{\natexlab{b}})Zhu, Zhang, He, Guo, Zeng, Qin, Zhang, and Gao]{zhu2023pointclip}
Xiangyang Zhu, Renrui Zhang, Bowei He, Ziyu Guo, Ziyao Zeng, Zipeng Qin, Shanghang Zhang, and Peng Gao.
\newblock Pointclip v2: Prompting clip and gpt for powerful 3d open-world learning.
\newblock In \emph{Proceedings of the IEEE/CVF International Conference on Computer Vision}, pages 2639--2650, 2023{\natexlab{b}}.

\bibitem[Zou et~al.(2025)Zou, Chen, and Yin]{zou2025}
Ying Zou, Zhe Chen, and Fuliang Yin.
\newblock High-order multi-scale attention and vertical discriminator enhanced clip for monocular depth estimation.
\newblock \emph{IEEE Transactions on Circuits and Systems for Video Technology}, pages 1--1, 2025.

\end{thebibliography}
}

\end{document}